\documentclass[preprint,5p,twocolumn,times, numbers]{elsarticle}

\usepackage{amssymb}
\usepackage{amsmath}

\usepackage{amsmath,amsfonts}
\usepackage{amsthm}
\usepackage{algorithmic}
\usepackage{algorithm}
\usepackage{array}
\usepackage[caption=false,font=normalsize,labelfont=sf,textfont=sf]{subfig}
\usepackage{textcomp}
\usepackage{stfloats}
\usepackage{url}
\usepackage{verbatim}
\usepackage{graphicx}
\usepackage{bm}
\usepackage{multirow}
\usepackage{colortbl}
\usepackage{booktabs}
\usepackage{hyperref}

\RequirePackage{amsthm}

\begin{document}

\begin{frontmatter}



\title{REAL: Representation Enhanced Analytic Learning for Exemplar-Free Class-Incremental Learning}


\author{Run He $^{a}$, Di Fang $^{b}$, Yizhu Chen $^{b}$, Kai Tong $^{a}$, Cen Chen $^{b}$, Yi Wang $^{c}$, Lap-pui Chau $^{c}$, Huiping Zhuang $^{a,*}$} 
\affiliation[a]{organization={Shien-Ming Wu School of Intelligent Engineering},
            addressline={South China University of Technology},
            city={GuangZhou},
            country={China}}
\affiliation[b]{organization={School of Future Technology},
            addressline={South China University of Technology},
            city={GuangZhou},
            country={China}}
\affiliation[c]{organization={Department of Electrical and Electronic Engineering},
            addressline={The Hong Kong Polytechnic University},
            state={Hong Kong SAR},
            country={China}}
\cortext[*]{Corresponding author: hpzhuang@scut.edu.cn}


\begin{abstract}

Exemplar-free class-incremental learning (EFCIL) aims to mitigate catastrophic forgetting in class-incremental learning (CIL) without {relying on historical training samples as exemplars}. Compared with { exemplar-based CIL that stores exemplars, EFCIL is more prone to forgetting}. {An emergent EFCIL} branch named Analytic Continual Learning (ACL) introduces a gradient-free paradigm {based on Recursive Least-Square for forgetting-resistant classifier training with a frozen backbone during CIL}. However,  {ACL currently} suffers from ineffective representations and {limited utilization of backbone knowledge}. {To address these challenges,} we propose a representation-enhanced analytic learning (REAL) {scheme}. {REAL improves the representation by constructing a dual-stream base pretraining stage followed by a representation-enhancing distillation process}. The dual-stream base pretraining combines self-supervised contrastive learning for general features {with} supervised learning for class-specific knowledge, {followed by representation-enhancing distillation to integrate both streams, improving representations for the subsequent CIL paradigm}. REAL {further introduces a feature fusion buffer to multi-layer backbone features, enabling richer feature extraction} for classifier training. Our {proposed} method can be incorporated into existing ACL techniques, {yielding superior performance}. Empirical results demonstrate that REAL achieves state-of-the-art performance on CIFAR-100, ImageNet-100, and ImageNet-1k benchmarks, outperforming {existing} exemplar-free methods and rivaling exemplar-based approaches.

\end{abstract}



\begin{keyword}


Class-Incremental Learning, Exemplar-Free, Analytic Learning, Representation Enhancement, Feature Fusion.
\end{keyword}

\end{frontmatter}



\section{Introduction}
\label{section:1}

Continual learning \citep{SASS2023KBS,SAGG2024KBS,KGP2025KBS}, also known as incremental learning, enables models to acquire knowledge from {sequentially arriving data in a phase-by-phase manner}. This process resembles the human learning, where new information is continuously assimilated into existing knowledge. {Similarly, continual learning} is crucial for the accumulation of machine intelligence in real-world scenarios. Class-incremental learning (CIL) \citep{LwF2018TPAMI,iCaRL2017_CVPR},  {a typical scenario of continual learning, has gained increasing popularity owing to} its ability to continuously adapt trained networks to {previously} unseen classes. {The approach recognizes} that {certain} data categories may be limited to certain locations or time slots, {thereby enabling} models to learn without resource-intensive joint retraining. 

\begin{figure}[h]
	\centering
	\includegraphics[width=1.0\linewidth]{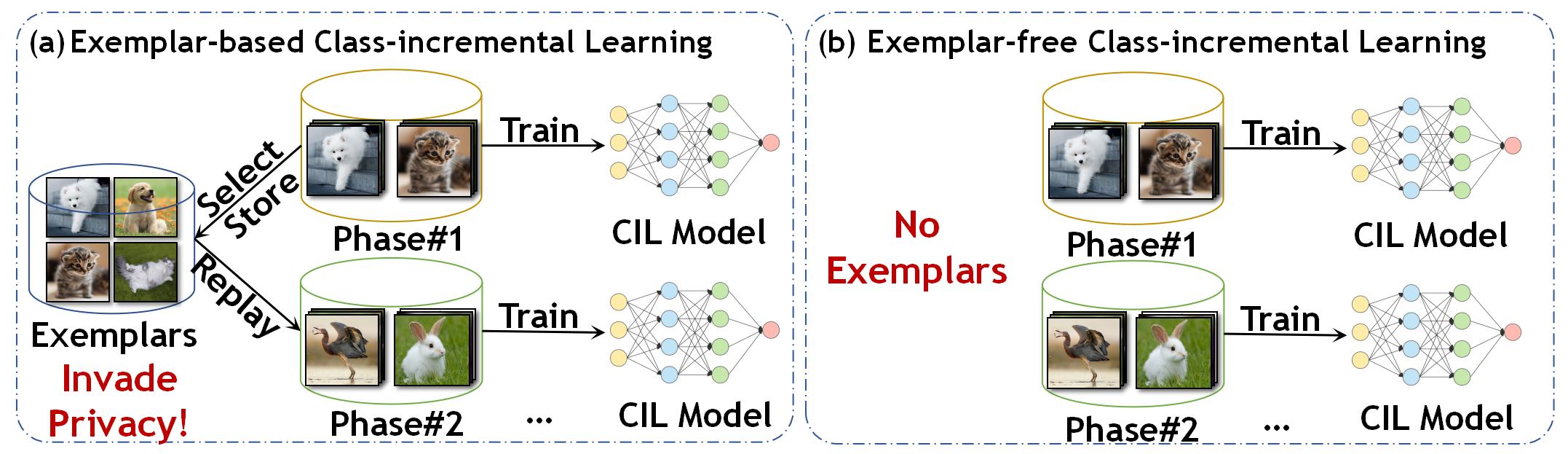}
	\caption{The difference between EBCIL and EFCIL. At each phase, EBCIL utilizes the stored exemplars selected from previous categories to train the model (as shown in (a)), while EFCIL does not keep exemplars and uses the categories within the training phase (i.e., in (b)). }
	\label{fig:intro}
\end{figure}

{Despite these advantages, CIL suffers from} \textit{catastrophic forgetting} \citep{cil_review2021NNs}, where models rapidly lose previously learned knowledge when acquiring new information. {Various solutions have been proposed to alleviate this issue. Among them}, exemplar-based CIL (EBCIL) has {demonstrated competitive results by storing historical samples as exemplars to help retain old knowledge while learning new data. However, this strategy introduces notable privacy concerns. By contrast, exemplar-free CIL (EFCIL) avoids storing previous samples, thus eliminating the privacy issue; however, it experiences more severe forgetting.} The difference between EBCIL and EFCIL is {illustrated} in Figure \ref{fig:intro}.

Recently, a new branch of EFCIL {termed} analytic continual learning (ACL) \citep{ACIL2022NeurIPS} {has effectively addressed} the catastrophic forgetting and has shown promising results while adhering to the exemplar-free constraint. ACL introduces a framework where a backbone is first trained using back-propagation (BP) in the base phase. {Subsequently,} a linear classifier is trained incrementally phase by phase with a recursive paradigm. {The recursive paradigm leverages analytic learning (AL) \citep{pil2001, brmp2021, cpnet2021} to train neural networks} with closed-form solutions and offers a \textit{weight-invariant property}. This property indicates that{ weights trained recursively in each phase are identical to those trained simultaneously on the entire data  \citep{ACIL2022NeurIPS,GKEAL2023CVPR}.} {ACIL incorporates this property into the CIL to achieve highly competitive results, particularly in large-phase scenarios (e.g., 50 phases) \citep{ACIL2022NeurIPS}}. The weight-invariant property also {maintains almost} the same results across different scenarios of phases. Moreover, {a fixed memory footprint for storing previous information in ACL enables ACL to operate} in extreme scenarios (e.g., learning one category in a phase \citep{AAAI2024DSAL}). 

However, existing ACL techniques have {two limitations}: (1) The feature representations are less effective on unseen categories after the base phase, {which limits} performance in future phases. Existing ACL methods {employ} supervised learning to acquire representation and {freeze} the backbone after the base phase, resulting in less discriminative representations across unseen data categories. (2)  {The classifier does not effectively utilize the knowledge captured by the backbone}. Existing ACL trains the classifier with the final output of the backbone. Features extracted by different layers of backbones can contain varying semantic information. For convolutional neural networks (CNNs) that {are} widely used  {in CIL tasks}, deep features are  {generally} more specific to the current task (i.e., classes in the current learning phase for CIL), whereas shallow features tend to be more transferable \citep{NIPS2014_shallow_deep_feature}. Learning solely with the final output may overlook potential improvements  {that can be achieved} with transferable shallow features. These limitations  {highlight} the  {importance of exploring} potential approaches to enhance representation learning and feature utilization of ACL. For representation enhancement, general knowledge learned without labels can  {benefit} incremental learning \citep{PRAKA2023ICCV} {. Additionally,} self-supervised contrastive learning (SSCL) \citep{SSL2020TPAMI} has shown effectiveness in learning general representations. SSCL learns general representations through contrastive proxy tasks, and these learned representations can be effectively adapted to downstream tasks.  {As} the model is initially trained in a base phase and then incrementally learns new tasks, general representations have the potential to enhance the ACL process. In terms of knowledge utilization, leveraging knowledge with different semantic information, particularly transferable features, {can }enhance classifier training in ACL. 

 {Herein}, we propose \textbf{R}epresentation \textbf{E}nhanced \textbf{A}nalytic \textbf{L}earning (REAL) for EFCIL to enhance the representation and knowledge utilization of ACL.  {Based on} the shared objectives of SSCL and CIL, REAL aims to improve the  {representation of the backbone} for unseen categories by integrating label-guided knowledge and general knowledge. Structurally, a dual-stream base pretraining (DS-BPT)  {is conducted for backbones } during the base phase to capture general knowledge {via SSCL and learn supervised feature distribution} via SL. Following DS-BPT, a representation-enhancing distillation (RED) process is employed to collaboratively merge both types of knowledge to further enhance the model. Consequently, a more robust backbone with improved representations is established for subsequent ACL. {Moreover, REAL enhances the utilization of backbone knowledge by introducing a feature fusion buffer (FFB)} to leverage features extracted from different layers of the CNN backbone. The FFB {employs a feature fusion process to construct informative feature vectors} that facilitate classifier training. REAL advances the ACL family by enhancing representation extraction without compromising fundamental benefits, such as the weight-invariant property{. These proposed} techniques are designed to seamlessly integrate into existing ACL frameworks. The key contributions of REAL are summarized as follows,

\begin{itemize}
{ 

    \item We introduce DS-BPT and RED to address the representation limitation in ACL. DS-BPT pretrains backbones in two streams using both SL and SSCL to acquire knowledge from both general and specific information. RED merges knowledge obtained from DS-BPT, constructing a backbone that encapsulates both types of knowledge, enabling the backbone to further refine its representation while retaining general knowledge.

    \item We propose FFB to overcome the challenge of knowledge utilization by constructing more informative features. FFB introduces mid-layer information by compressing information of different channels to a layer-wise feature map. Subsequently, the compressed feature maps extracted from different layers of the backbone are fused to create feature vectors, combining transferable and specific features for classifier learning.

    \item The components introduced in REAL can be seamlessly integrated into existing ACL methods, which makes REAL a versatile technique to enhance existing ACL approaches.

    \item Experimental results on various benchmark datasets demonstrate that REAL achieves remarkable performance, outperforming state-of-the-art EFCIL methods and several EBCIL methods. 
    }
\end{itemize}






The remaining contents of this paper are organized as follows. Section \ref{section:2} reviews the related works. Our method, namely REAL, is depicted in Section \ref{section:4}. In Section \ref{section:5}, experiments are conducted to validate and analyze the performance of REAL. Section \ref{section:6} concludes this paper.

\section{Related Works}
\label{section:2}
\noindent In this section, we provide a comprehensive review of CIL techniques covering both EBCIL and EFCIL methods. { Additionally, SSCL, which is closely related to the proposed REAL, is also discussed.}

\subsection{Exemplar-Based CIL} EBCIL, {first introduced by iCaRL \citep{iCaRL2017_CVPR}, reinforces models' memory by replaying historical exemplars via a herding mechanism for selecting exemplars}. The exemplars are a fraction of encountered data stored in a fixed memory. Such a replay mechanism has gained popularity with various {subsequent attempts \citep{TPCIL2020ECCV, GeoDL2021CVPR}. Various studies have leveraged knowledge distillation (KD) to gain better replay}. For instance, LUCIR \citep{LUCIR2019_CVPR} adjusts the feature to a sphere surface with cosine similarity and uses KD to maintain the orientations of features. PODNet \citep{podnet2020ECCV} implements an efficient spatial-based distillation loss to reduce forgetting. FOSTER \citep{FOSTER2022ECCV} employs a two-stage learning paradigm with KD, expanding and then reducing the network to the original size. {In addition, numerous studies have investigated the management of memory.} {MIR \citep{MIR2019NeurIPS} constructs a better exemplar set within limited memory by selecting} samples that have a large impact on loss. GDumb \citep{GDumb2020ECCV} greedily stores samples while keeping the classes balanced. RMM \citep{RMM2021NeuriPS} leverages reinforcement learning to dynamically manage memory for exemplars. The scope of stability-plasticity balance during replaying {was} also studied. AANets \citep{AANet_2021_CVPR} integrates a stable block and a plastic block to balance stability and plasticity{. Additionally,} online hyperparameter optimization \citep{Online2023AAAI} adaptively optimizes the stability-plasticity balance without prior knowledge. Although the EBCIL provides generally good results, it invades privacy and requires {additional memory to store exemplars.}  

\subsection{Exemplar-Free CIL} Compared with EBCIL, EFCIL addresses catastrophic forgetting without the {using exemplars to alleviate the privacy problem}. It can be categorized into regularization-based CIL \citep{MAS2018ECCV,CRNet2023TPAMI,PC2018ICML}, prototype-based CIL \citep{SDC2020CVPR,Fusion2022CVPR,POLO2023MM},
and the recently proposed ACL  \citep{ACIL2022NeurIPS,AAAI2024DSAL,GKEAL2023CVPR}. 

\textbf{Regularization-based CIL} imposes additional constraints on network activations or key parameters to mitigate forgetting. KD is introduced as {a typical method to construct the constraint between old and new models.} For example, LwF \citep{LwF2018TPAMI} prevents activation changes between old and new networks via KD, and LwM \citep{LWM2019CVPR} introduces an attention distillation loss. Other approaches {set constraints related to important weights of previous tasks, and several researches have studied how to model the importance}. For example,  EWC \citep{EWC2017nas} captures prior importance using a diagonally approximated Fisher information matrix. Building upon  EWC, online EWC  \citep{PC2018ICML} reduces the accumulations of Fisher matrices in EWC. SI \citep{SI2017ICML} calculates the impact of parameter change on loss via the path integral to determine important parameters. RWalk \citep{RWalk2018ECCV} combines the Fisher matrix in EWC and path integral in SI to compute the importance. {Regularization-based CIL} respects privacy without storing exemplars, but the constraint imposed on network updates prevents the model from learning new tasks.

\textbf{Prototype-based CIL} mitigates forgetting by storing prototypes to maintain decision boundaries across old and new tasks. The prototypes, typically the feature means of specific classes, {are incorporated into classifier training to supply previous information}. Multiple studies have investigated how to simulate past feature distributions from prototypes. For instance, PASS \citep{PASS2021CVPR} augments the prototypes by {adding Gaussian noise to balance between old and new categories}. SSRE \citep{SSRE2022CVPR} introduces a prototype selection mechanism with up-sampling of prototypes to reduce feature confusion. FeTriL \citep{FeTrIL2023WACV} introduces a mechanism to generate pseudo-features of old classes by estimating the difference of feature distributions from prototypes. However, these methods do not consider {the fact that, old prototypes} may be inaccurate due to the evolution of models during incremental learning.  {PRAKA \citep{PRAKA2023ICCV} addresses this problem by providing a prototype reminiscence mechanism} that estimates the change of old prototypes. Moreover, NAPA-VQ \citep{NAPA-VQ2023ICCV} utilizes the {neighborhood} information in feature space to hold more compact decision boundaries between classes. As a prevalent branch recently proposed, the prototype-based CIL gains prominence in the EFCIL realm and achieves state-of-the-art performance. However, it favors stability with limited plasticity, and the storing of prototypes can be a large extra cost for large-class tasks.  


\textbf{ACL} is a recently developed branch of EFCIL that aims to find closed-form solutions {for} all seen data.{AL \citep{pil2001,brmp2021,cpnet2021} utilizes} least squares (LS) to yield closed-form solutions during the training of neural networks. ACIL \citep{ACIL2022NeurIPS} first introduces AL to the CIL realm. It is developed by reformulating the CIL procedure into a recursive LS form {thereby equating the CIL with} joint training in the linear classifier. The subsequent GKEAL \citep{GKEAL2023CVPR} specializes in few-shot CIL settings by adopting a Gaussian kernel process that excels in data-scarce scenarios. DS-AL \citep{AAAI2024DSAL} introduces a compensation stream to address the fitting {problem} of the linear classifier in ACIL. The ACL family {can} tackle various scenarios, including online CIL \citep{FOAL2024NeurIPS}, imbalanced CIL \citep{AIR2024, AEF-OCLTVT2025}, generalized CIL \citep{GACL2024NeurIPS}, and multi-modal tasks \citep{MMAL2024MM, RAIL2024NeurIPS}.  While ACL demonstrates substantial performance, it relies on the {performance of the backbone, which is trained} solely on the base dataset in a supervised manner. This approach {may result in} inadequate representations of unseen data without future label guidance. Moreover, training {solely} with final features of the backbone diminishes the utilization of backbone knowledge, potentially impacting {of ACL. This study introduces REAL to enhance ACL and mitigate these limitations.}

\subsection{Self-Supervised Contrastive Learning} SSCL \citep{SSL2020TPAMI} enables label-free learning via a contrastive manner. It encourages features of similar samples to get close and forces dissimilar ones to {be} apart. {Generally}, Siamese networks are constructed for contrastive learning. For instance, SimCLR \citep{SimCLR2020ICML} utilizes a symmetric Siamese network for contrastive training and MOCO \citep{MOCO2020CVPR} mitigates the influence of extreme samples through momentum updates. However, they {require} negative samples (samples in different classes) and large a batch size to train a good model. {By contrast}, BYOL \citep{BYOL2020NeurIPS} uses only positive samples by introducing auxiliary structures trained with momentum.  Built upon BYOL, SimSiam \citep{SimSiam2021CVPR} replaces the momentum mechanism with gradient stopping and {achieves} considerable performance with a small batch size. The models trained {via} SSCL learn to extract general representations and can be effectively transferred \citep{SSL2023TKDE}. We utilize this technique to obtain the general base representations. {Several studies has used} contrastive learning to improve the performance of CIL \citep{Co2L2021ICCV, CaSSLe2022CVPR, SSLEFCIL2024WACV}. However, these approaches either restrict the use of contrastive learning to EBCIL \citep{Co2L2021ICCV, CaSSLe2022CVPR}, or {rely solely on} the knowledge acquired through SSCL during the incremental phases \citep{SSLEFCIL2024WACV}. {By} contrast,  {REAL is the first method} to integrate knowledge obtained from both SSCL and SL to enhance incremental learning within the EFCIL framework.



\begin{figure*}[t]
	\centering
	\includegraphics[width=1\linewidth]{REAL-flow_New_V6.png}
	{\caption{REAL focuses on enhancing the representation and backbone knowledge utilization during training and comprises three parts: (a) Dual-stream base pretraining (DS-BPT), (b) representation enhancing distillation (RED), and (c) analytic learning with feature fusion. Base knowledge is obtained by pretraining the CNN backbones with two streams. One stream acquires general base knowledge (GBK) via self-supervised contrastive learning (SSCL). The other learns supervised feature distribution (SFD) via supervised learning (i.e., in (a)). During RED, the backbone with SFD is frozen and transfers knowledge to the model with GBK via knowledge distillation (i.e., in (b)). Following base training, the enhanced backbone (i.e., backbone with GBK after RED) is embedded in the analytic learning agenda with a feature fusion buffer (i.e., in (c)).}}
	\label{fig:mainflow}
\end{figure*}

\section{Proposed Method}
\label{section:4}
\noindent {This section presents the details of the proposed REAL framework. The paradigm comprises three parts}: Dual-stream Base Pretraining (DS-BPT), Representation Enhancing Distillation (RED) and Analytic Learning with Feature fusion (ALFF). An overview of REAL is depicted in Figure \ref{fig:mainflow}. 

\subsection{Dual-Stream Base Pretraining}
Prior to further description, some definitions related to the CIL are presented. For CIL, the training data are given {in a phase-by-phase manner} with disjoint data categories. Suppose the CIL paradigm contains $K+1$ phases, at each phase (e.g., phase $k$, $k=0,\dots,K$), training data of disjoint classes $\mathcal{D}_{k}=\{\mathcal{X}_{k,i}, y_{k,i}\}_{i=1}^{N_{k}}$, where $\mathcal{X}_{k,i} \in \mathbb{R}^{ c\times w \times h}$ {represents} the $i$-th input image in task $k$, $y_{k,i}$ denotes the label of the $i$-th input data, and $N_k$ {indicates} the total number of samples in task t. {In particular}, $\mathcal{D}_{0}^{\text{train}}=\{\mathcal{X}_{0,i}^{\text{train}}, y_{0,i}^{\text{train}}\}_{i=1}^{N_{0}}$ represents the base training set and $\mathcal{D}_{0:k}$ {represents} the joint dataset from phase $0$ to phase $k$. The objective of CIL at phase $k$ is to train the model given $\mathcal{D}_{k}^{\text{train}}$ and test it on $\mathcal{D}_{0:k}^{\text{test}}$ .

ACL suffers from less effective representation after the base phase{. We} propose dual-stream base pretraining at the base phase to capture the both general base knowledge and specific supervised feature distribution (label-guided knowledge). 
Technically, the network is pretrained in streams of supervised learning and SSCL to obtain the base knowledge during base phase before a subsequent enhancement. Here we use a CNN as the backbone.

\textbf{General Base Knowledge Acquisition.} The {GBK acquisition stream trains a backbone with GBK via the SSCL which decouples training from labels}. We adopt SimSiam \citep{SimSiam2021CVPR} as the SSCL architecture, which comprises a Siamese network of CNN backbone and formulates contrastive proxy tasks between different views (i.e., different augmentations of the same image).   

Auxiliary structures, namely projector and predictor, are incorporated to assist SSCL training as extra mappings. The projector and predictor are multi-layer perceptrons{(MLPs)} that provide a diversified feature mapping {benefiting} the SSCL process. Suppose the stacked $N_{k}$-sample input tensor at phase 0 is $\bm{X}^{\text{train}}_{0}$, and the random data augmentation operation is denoted as $f_{\text{RA}}$, two views of the input image tensors after data augmentation can be represented as $\bm{X}^{\text{aug}_{1}} =f_{\text{RA}}(\bm{X}^{\text{train}}_{0})$ and $\bm{X}^{\text{aug}_2}=f_{\text{RA}}(\bm{X}^{\text{train}}_{0})$. Let $\bm{W}_{\text{proj}}$, $\bm{W}_{\text{pred}}$, and $\bm{W}_{\text{CNN}}^{\text{GBK}}$ represent the weights of the projector, predictor, and the CNN backbone for GBK acquisition respectively. Given two views ($\bm{X}^{\text{aug}_1}$, $\bm{X}^{\text{aug}_2}$) of input  $\bm{X}\in\mathbb{R}^{c \times w\times h }$, the outputs are,
\begin{align}
	\bm{X}^{\text{proj}_1} &=f_{\text{proj}}(f_{\text{flat}}(f_{\text{CNN}}(\bm{X}^{\text{aug}_1}, \bm{W}_{\text{CNN}}^{\text{GBK}})\bm{W}_{\text{proj}})), \\
	\bm{X}^{\text{proj}_2} &=f_{\text{proj}}(f_{\text{flat}}(f_{\text{CNN}}(\bm{X}^{\text{aug}_2}, \bm{W}_{\text{CNN}}^{\text{GBK}})\bm{W}_{\text{proj}})), \\
	\bm{X}^{\text{pred}_1} &= f_{\text{pred}}(\bm{X}^{\text{proj}_1},\bm{W}_{\text{pred}} ), \\
\bm{X}^{\text{pred}_2} &= f_{\text{pred}}(\bm{X}^{\text{proj}_2},\bm{W}_{\text{pred}} ),
\end{align} 
where $f_{\text{CNN}}$ {indicates passing the image through the CNN backbone}, $f_{\text{flat}}$ {represents} a flattening operator, {and $f_{\text{proj}}$ and $f_{\text{pred}}$ are operations of projector and predictor respectively}.

{We then} optimize the negative cosine similarity of these features. For two tensors $\bm{Z}_1$ and $\bm{Z}_2 \in\mathbb{R}^{N\times d} $, the negative cosine similarity can be defined as
\begin{align}
	\mathcal{L}_{\text{cos}}(\bm{Z}_1, \bm{Z}_2) = -\sum_{i=1}^{N} \frac{\bm{z}_{1,i}\cdot \bm{z}_{2,i}}{\left\lVert\bm{z}_{1,i}\right\rVert_{2} {\left\lVert\bm{z}_{2,i}\right\rVert_{2}}}, 
\end{align} 
where $\bm{z}_{j,i}$ {represents} a row vector of $\bm{Z}_j$ ($i = 1, 2, \dots , N$; $j = 1, 2$), $\cdot$ {refers to the} dot product operation, and $\lVert\cdot\rVert_{2}$ is the L2-norm.
The loss function of GBK acquisition{is} 
\begin{align}\label{L1}
	\mathcal{L}_{\text{GBK}} = \frac{1}{2}\mathcal{L}_{\text{cos}}(\bm{X}^{\text{proj}_1}, \bm{X}^{\text{pred}_2}) + \frac{1}{2}\mathcal{L}_{\text{cos}}(\bm{X}^{\text{proj}_2}, \bm{X}^{\text{pred}_1}),
\end{align} 
{and we can use BP to minimize $\mathcal{L}_{\text{GBK}}$ to obtain a finely trained model. }


\textbf{Learning Supervised Feature Distribution.} While we can capture the {GBK via} an SSCL process, the information derived from labels remains valuable. Supervised knowledge can serve as a beneficial complement, particularly when conducting CIL in a supervised manner. Here, we employ a learning stream to acquire specific knowledge, specifically focusing on supervised feature distribution through supervised learning. Suppose{$\bm{W}_{\text{CNN}}^{\text{SFD}}$ and  $\bm{W}^{\text{SFD}}_{\text{FCN}}$ represent} the weights of the CNN backbone and fully-connected network (FCN), the output can be 
\begin{align}
	\bm{\hat Y}^{\text{SFD}} = f_{\text{softmax}}(f_{\text{flat}}(f_{\text{CNN}}(\bm{X}_{0}^{\text{train}}, \bm{W}_{\text{CNN}}^{\text{SFD}}))\bm{W}_{\text{FCN}}^{\text{SFD}}).
\end{align}
With labels $\bm{Y}_{0}^{\text{train}}$, we can train the model via BP with the loss function of cross entropy. {Following} training, the backbone can generate the feature with certain distribution guided by labels. {DS-BPT encodes both GBK and SFD into two backbones and then utilizes them in the subsequent procedures.} 

\subsection{Representation Enhancing Distillation}
{Both general knowledge (GBK) and specific knowledge (SFD) are obtained after DS-BPT.} Here, we introduce the RED process to leverage both types of knowledge to enhance the representation of the backbone.

We enhance the representation by transferring supervised knowledge through the extracted feature vectors {using a KD process.} The extracted feature vectors from two backbones can be {obtained} as 
\begin{align}
\bm{X}^{\text{GBK}} = f_{\text{flat}}(f_{\text{CNN}}(\bm{X}^{\text{train}}_{0}, \bm{W}_{\text{CNN}}^{\text{GBK}})), \\
\bm{X}^{\text{SFD}} = f_{\text{flat}}(f_{\text{CNN}}(\bm{X}^{\text{train}}_{0}, \bm{W}_{\text{CNN}}^{\text{SFD}})).
\end{align}

During the KD process, the backbone pretrained in {the SFD} is frozen to provide knowledge of feature distribution, while the backbone pretrained in the GBK stream assimilates the knowledge. This is accomplished by aligning the output features extracted by the GBK backbone with those extracted by the SFD model by minimizing the difference them. {In particular, we utilize cosine similarity, i.e.,}
\begin{align}
	\mathcal{L}_{\text{feature}} = -\frac{1}{N}\sum_{i=1}^{N} \frac{\bm{x}^{\text{SFD}}_{i} \cdot \bm{x}^{\text{GBK}}_{i}}{\left\lVert \bm{x}^{\text{SFD}}_{i}\right\rVert_{2} \left\lVert\bm{x}^{\text{GBK}}_{i}\right\rVert_{2}}, 
\end{align}
where $\bm{x}_{{i}}^{\text{SFD}}$ and $\bm{x}_{{i}}^{\text{GBK}}$ {represent} $i$-th row vectors of $\bm{X}^{\text{SFD}}$ and $\bm{X}^{\text{GBK}}$ respectively. 
The labels can serve as a valuable source of supervision for enhancing representations. We incorporate additional cross-entropy loss for classification to utilize the direct guidance offered by the labels. {We introduce a linear classifier after the backbone to correlate the representations extracted by the backbones with the labels.} The output of the model following this addition is denoted as 
\begin{align}
	\bm{\hat Y}^{\text{GBK}} = f_{\text{softmax}}(f_{\text{flat}}(f_{\text{CNN}}(\bm{X}_{0}^{\text{train}}, \bm{W}_{\text{CNN}}^{\text{GBK}}))\bm{W}_{\text{FCN}}^{\text{GBK}}).
\end{align}
The cross-entropy loss is {given by},
\begin{align}
	\mathcal{L}_{\text{label}} = -\frac{1}{N}\sum_{i=1}^{N}\log{\bm{\hat y}^{\text{GBK}}_{i}}\bm{y}_{0,i}^{\text{train}}, 
\end{align}
where $\bm{\hat y}_{i}^{\text{GBK}}$ {represents} the $i$-th row vectors of $\bm{\hat Y}^{\text{GBK}}$.
Finally, we can have the integrated loss of {RED}
\begin{align}
	\mathcal{L}_{\text{RED}} = \lambda \mathcal{L}_{\text{feature}} + (1-\lambda) \mathcal{L}_{\text{label}},
\end{align}
where $\lambda$ is the parameter for balancing the contribution of the teacher model and the labels. 

Thus, {we can merge the knowledge from two sources with extra knowledge of labels injected. Following} the RED, the trained CNN backbone is frozen and serves as the backbone.

\textbf{Why RED Needs Two Sources of Supervision?} RED is designed to improve the representation of the backbone trained in the base phase by integrating supervised information with general knowledge. Depending exclusively on labels can result in direct adaptation to base phase labels, potentially causing forgetting in subsequent learning phases. Relying solely on the feature distribution from the SFD backbone may not be entirely effective {because} it could potentially mislead the GBK backbone. Hence, a fusion of both knowledge sources is crucial, a pattern supported by our experimental findings (refer to Section \ref{sec:hyper}).
 
\textbf{Enhancing Future Representations with a Frozen Backbone.} Existing ACL methods {are constrained by limited representations.} Backbones trained through SL struggle to {effectively} capture the distinctive features of forthcoming categories, thereby restricting the performance of linear classifiers. The general knowledge learned without the labels {is} highly transferable in the future phases and our proposed representation enhancement incorporates general knowledge and information under supervision. Although the backbone is frozen after RED, it yields more distinguishable representations, thereby facilitating classification in CIL {. Although training the backbone during CIL can directly improve future representations, it can easily result in} catastrophic forgetting on the old phases and yields a bad performance. Freezing the backbone presents an effective strategy to mitigate forgetting, inspiring the development of various CIL techniques based on a frozen backbone \citep{FeTrIL2023WACV, FeCAM2023NeurIPS} including ACL methods. {REAL aligns with these CIL techniques, offering improved representations for future tasks.}



\subsection{Analytic Classifier with Feature Fusion}

{Upon} obtaining the enhanced backbone via RED, we employ {AL} to achieve weight-invariance in classifier during the incremental learning process. {Following} RED, the backbone is frozen, and the classifier is replaced with an analytic classifier comprising a buffer layer and {a FCN}. The buffer layer, a common component in existing ACL techniques, {introduces} a non-linear mapping that projects features into a higher-dimensional space \citep{ACIL2022NeurIPS}. According to Cover's theorem, non-linear mapping to higher-dimensional space enhances feature separability \citep{Cover1965TEC}. However, existing ACL techniques utilize the buffer solely {to process} the final output of the backbone, potentially neglecting the transferable knowledge embedded in the backbone. {We address this limitation by introducing the Feature Fusion Buffer (FFB) to enhance the exploitation of knowledge acquired in the base phase}.

\textbf{Feature Fusion Buffer.} Feature maps attained by different blocks of a CNN {can} indicate different semantic information, where the shallow features {are} more transferable and the deep features can be more specific \citep{NIPS2014_shallow_deep_feature}. {We propose a feature fusion process to utilize all these features}. At phase 0, suppose the output feature map of $i$-th block of the CNN backbone is denoted as $\bm{X}_{0}^{i} \in \mathbb{R}^{N_{k}\times c_i\times w_i \times h_i}$, then the output after the FFB $\bm{X}_{0}^{\text{(FFB)}}$ can be obtained {using the following equations}
\begin{align}\label{eq_buffer}
	\bm{X}_{0}^{\text{(FFB)}} = \frac{1}{N_b} \sum_{i=1}^{N_b} \sigma(f_{\text{ap}}(\bm{X}_{0}^{i})\bm{W}_{i,\textup{B}})),
\end{align}
where $f_{\text{ap}}$ {denotes} the operation of the average pooling on the dimension of $w_i$ and $h_i$, $\bm{W}_{i,\textup{B}} \in \mathbb{R}^{c_i\times d_{B}}$ {represents} a randomly initialized matrix for projecting to $d_{B}$ dimensional space, $N_b$ {indicates} the total number of layers in CNN, and $\sigma$ {represents} a non-linear activation function (we use ReLU in this paper). {This average pooling summarizes the channel-wise information within each block, providing diversified information for the classifier training. 
}

The benefit of FFB can be summarized in three folds: (1) FFB adheres to the benefit of projecting features into a high-dimensional space, {enhancing} the separability of features based on Cover's theorem. (2) FFB provides more informative {features that contain} transferable knowledge, thus benefits the incremental learning.  (3) FFB is training-free, making it an easy yet effective technique to enhance the classifier training in ACL.

{
The FFB shares similar idea of utilizing multi-layer features in F-OAL \citep{FOAL2024NeurIPS}, however, FFB further proposes a channel-wise information condensation of each layer. Based on Vision Transformer \citep{ViT}, the feature fusion in F-OAL simply takes the average of class token in each layer. FFB in REAL, by contrast, utilizes an extra pooling to compress and summarize channel-wise information to provide more diversified features. The feature maps of each channel provide different perspectives of the input image \citep{Comp2021ICML} and previous study demonstrates the effectiveness of channel-wise information \citep{podnet2020ECCV}, making FFB an effective module in the continual learning of CNNs. 
}

\textbf{Classifier Training via Analytic Learning.}  {Following} the feature fusion, the features are provided to the subsequent FCN layer and the {AL} is used to train the FCN. Following the ACL process \citep{ACIL2022NeurIPS}, the weight of the FCN layer can be obtained at the base phase by optimizing the following equation
\begin{align}\label{eq_L2}
	\underset{\bm{W}_{\text{A}}}{\text{argmin}} \quad \left\lVert\bm{Y}_{0}^{\text{train}} - \bm{X}_{0}^{\text{(FFB)}}\bm{W}_{\text{A}}^{(0)}\right\rVert_{\text{F}}^{2} + {\gamma} \left\lVert\bm{W}_{\text{A}}^{(0)}\right\rVert_{\text{F}}^{2},
\end{align}
where $\lVert\cdot\rVert_{\text{F}}^2$ is Frobenius-norm, $\gamma$ is the regularization factor, $\bm{W}_{\text{A}}^{(0)}$ {denotes} the weight of FCN layer at the base phase, and $\bm{Y}_{0}^{\text{train}}$ {denotes} the stacked one-hot label tensor.

The optimal solution to \eqref{eq_L2} is 
\begin{align}\label{eq_ls_w_base}
	\bm{\hat W}_{\text{A}}^{(0)} = (\bm{X}_{0}^{\text{(FFB)T}}\bm{X}_{0}^{\text{(FFB)}}+\gamma \bm{I})^{-1}\bm{X}_{0}^{\text{(FFB)T}}\bm{Y}_{0}^{\text{train}},
\end{align}
where $\bm{\hat W}_{\text{A}}^{(0)}$ indicates the estimated weight of the FCN layer at phase 0, and $\cdot^{\text{T}}$ {represents} the matrix transpose operator.

For the learning in subsequent phases, let 
\begin{equation}\label{eq_emb_label}
 \bm{X}_{0:k}^{\text{(FFB)}} = \begin{bmatrix}
		\bm{X}_{0}^{\text{(FFB)}}\\
        \bm{X}_{1}^{\text{(FFB)}}\\
        \vdots\\
        \bm{X}_{k}^{\text{(FFB)}}
	\end{bmatrix},  \quad
	\bm{Y}_{0:k} = \begin{bmatrix}
		\bm{Y}_{0}^{\text{train}} \\
		\bm{Y}_{1}^{\text{train}} \\
		\vdots \\
		\bm{Y}_{k}^{\text{train}}
	\end{bmatrix}
\end{equation}

At phase k, the learning problem of all seen data $\mathcal{D}_{0:k}^{\text{train}}$ can be extended from \eqref{eq_L2} to
\begin{align}\label{eq_ls_k-1}
		\underset{\bm{W}_{\text{A}}^{(k)}}{\text{argmin}} \quad \left\lVert	\bm{Y}_{0:k}^{\text{train}} -
		\bm{X}_{0:k}^{\text{(FFB)}} \bm{W}_{\text{A}}^{(k)}\right\rVert_{\text{F}}^{2} + {\gamma} \left\lVert\bm{W}_{\text{A}}^{(k)}\right\rVert_{\text{F}}^{2}.
\end{align}

The solution of \eqref{eq_ls_k-1} is 
\begin{align}\label{eq_ls_w_k-1}
    \bm{\hat W}_{\text{A}}^{(k)} &= (\bm{X}_{0:k}^{\text{(FFB)T}}\bm{X}_{0:k}^{\text{(FFB)}}+\gamma \bm{I})^{-1}\bm{X}_{0:k}^{\text{(FFB)T}}\bm{Y}_{0:k}^{\text{train}}. 
\end{align}

The goal for CIL is to {sequentially} learn new tasks on $\mathcal{D}_{k}^{\text{train}}$ given a network trained on $\mathcal{D}_{0:k-1}^{\text{train}}$. Here, to cancel reliance on previous data in \eqref{eq_ls_k-1}, let $\bm{R}_{k} =(\bm{X}_{0:k}^{\text{(FFB)T}}\bm{X}_{0:k}^{\text{(FFB)}}+\gamma\bm{I})$ and $\bm{Q}_{k} =\bm{X}_{0:k}^{\text{(FFB)T}}\bm{Y}_{i}^{\text{train}}$, the problem can be solved in the following transformation, 

\begin{align}\nonumber
    \bm{\hat W}_{\text{A}}^{(k)} &= \bm{R}_{k}^{-1}\bm{Q}_{k}. \\ \nonumber
    &= (\sum_{i=0}^{k}\bm{X}_{i}^{\text{(FFB)T}}\bm{X}_{i}^{\text{(FFB)}}+\gamma \bm{I})^{-1}\sum_{i=0}^{k}\bm{X}_{i}^{\text{(FFB)T}}\bm{Y}_{i}^{\text{train}} \\  \label{eq_rq_recursive}
    &= (\bm{R}_{k-1}+\bm{X}_{k}^{\text{(FFB)T}}\bm{X}_{k}^{\text{(FFB)}})^{-1} (\bm{Q}_{k-1} + \bm{X}_{k}^{\text{(FFB)T}}\bm{Y}_{k}^{\text{train}}),
\end{align}
where $\bm{X}_{i}^{\text{(FFB)}}$ denotes the feature tensor after FFB at phase i. 

Equation \eqref{eq_rq_recursive} indicates {that} the results of joint training in \eqref{eq_ls_k-1} can be reproduced by {sequentially performing recursively training on $\mathcal{D}_{k}^{\text{train}}$}. That is, the information of previous phases 0:k-1 can be encoded in $\bm{R}_{k-1}$ and $\bm{Q}_{k-1}$, and the new information can be included via addition, {as expressed in} Equation \eqref{eq_rq_recursive}. This implies the property that CIL is equalized to joint training with both present and historical data, i.e., the \textit{weight-invariant property} \citep{ACIL2022NeurIPS} shared in the ACL family. This pattern ensures the non-forgetting in ACL and {yields} high performance. Our proposed REAL model also adheres to this property, exhibiting strong performance in mitigating forgetting. Additionally, by leveraging RED and FFB, {the proposed REAL model can further enhance the efficacy of training features within the FCN, significantly improving performance compared to ACL techniques.}

\textbf{Incorporate REAL into DS-AL.} Here, we {demonstrate how  REAL can be effectively integrated into another ACL technique, namely DS-AL \citep{AAAI2024DSAL}}, which enhances ACIL by introducing a compensation stream to learn residues. Following DS-AL, we incorporate a compensation stream into REAL and refer to this modified version as REAL-DS. REAL-DS maintains the same paradigm of backbone training to utilize RED and the modification lies on the ALFF. 

At phase k, {upon learning from the labels} (i.e., the main stream), the training of the compensation stream to fit the residuals commences. Let ${\bm{\tilde{Y}}}_{k}$ {denote} the residue of final output after conducting the main stream, {which is} calculated as 
\begin{align}\label{eq_res}
	{\bm{\tilde{Y}}}_{k} = {\bm{Y}}_{k}^{\text{train}} - \bm{X}_{k}^{(\text{FFB})}\bm{\hat W}_{\text{A}}^{(k)} [:,:-d_{yk}],
\end{align}
where $ \bm{X}_{k}^{(\text{FFB})}\bm{\hat W}_{\text{A}}^{(k)} [:,:-d_{yk}]$ represents  the logits of current learning classes and $d_{yk}$ {indicates} the number of the classes in phase k. Equation \eqref{eq_res} computes the local residue for the current learning phase, and a compensation stream is employed to rectify this error. The objective of the compensation stream is to 
\begin{align}\label{eq_ls_k-1_C}
		\underset{\bm{W}_{\text{A,C}}^{(k)}}{\text{argmin}} \quad \left\lVert	\bm{\tilde{Y}}_{k} -
		\bm{X}_{0:k,C}^{\text{(FFB)}} \bm{W}_{\text{A,C}}^{(k)}\right\rVert_{\text{F}}^{2} + {\gamma} \left\lVert\bm{W}_{\text{A,C}}^{(k)}\right\rVert_{\text{F}}^{2},
\end{align}
where $\bm{X}_{C,0:k}^{\text{(FFB)}} = \frac{1}{N_b} \sum_{i=1}^{N_b} \sigma_{C}(f_{\text{ap}}(\bm{X}_{0:k}^{i})\bm{W}_{i,\textup{B}}))$ {represents} the input of the compensation stream with {activation function $\sigma_C$, which is used to alternate the input of the compensation stream} to improve the fitting ability \citep{AAAI2024DSAL}. {The} solution to Equation \eqref{eq_ls_k-1_C} is 
\begin{align}\label{eq_ls_w_k-1_C}
    \bm{\hat W}_{\text{A,C}}^{(k)} &= (\bm{R}_{k-1,C}+\bm{X}_{k,C}^{\text{(FFB)T}}\bm{X}_{k,C}^{\text{(FFB)}})^{-1} (\bm{Q}_{k-1,C} + \bm{X}_{k,C}^{\text{(FFB)T}}\bm{\tilde{Y}}_{k}), \\ \label{eq_ls_rq_k-1_C}
    \bm{R}_{k-1,C} &=(\bm{X}_{0:k-1,C}^{\text{(FFB)T}}\bm{X}_{0:k-1,C}^{\text{(FFB)}}+\gamma\bm{I}), \quad
    \bm{Q}_{k-1,C} =\bm{X}_{0:k-1,C}^{\text{(FFB)T}}\bm{\tilde{Y}}_{0:k-1},
\end{align}
which {completes the learning of the compensation stream}. The inference with compensation stream is produced by combining both streams as $\bm{\hat Y}_{k} = \bm{X}_{k}^{(\text{FFB})}\bm{\hat W}_{\text{A}}^{(k)} + \mathcal{C}\bm{X}_{k,C}^{(\text{FFB})}\bm{\hat W}_{\text{A,C}}^{(k)}$, where $C$ is the factor to control the contribution of compensation stream. 

{Accordingly}, our REAL can be an seamlessly integrate into DS-AL. This  demonstrates that, REAL can function as a plug-and-play module capable of improving established ACL techniques via enhanced representations and informative features.

\section{Experiments}
\label{section:5}
\noindent {This section discusses comprehensive experiments on various benchmark datasets to validate REAL}. First, we compare the proposed REAL with various EFCIL and EBCIL methods. Subsequently, the analysis on hyperparameters, representation enhancement and ablation study of knowledge sources in REAL are provided. 

\subsection{Experimental Setup}
\textbf{Datasets.} We evaluate the performance of REAL together with existing ACL methods on  CIFAR-100 \citep{CIFAR2009}, ImageNet-100 \citep{podnet2020ECCV}, and ImageNet-1k  \citep{ImageNet2015IJCV}. CIFAR-100 contains 100 classes of 32 $\times$ 32 color images with each class  {containing} 500 images for training and 100 images for testing respectively. ImageNet-1k has 1.3 million images for training and 50,000 images for testing within 1000 classes. ImageNet-100, as defined in \citep{podnet2020ECCV}, is constructed by selecting 100 specific classes from ImageNet-1k. 

\textbf{CIL Protocol.} For CIL evaluation, we follow the protocol adopted in various CIL  {studies} \citep{podnet2020ECCV, Mnemonics_2020_CVPR, PASS2021CVPR}, which  {involves partitioning the} dataset into a base training set (i.e., phase 0) and a sequence of incremental sets. The base dataset contains half of the full data classes. The network is first trained on the base dataset. Subsequently, the network gradually learns the remaining classes evenly for $K$ phases. In the experiment, we report the results for phase $K=5,10,25$, and $50$ on CIFAR-100 and ImageNet-100.  {Additionally, we} validate our REAL on ImageNet-1k with $K= 5,10,$ and $25$.

\textbf{Implementation Details.} The architecture in the experiment is ResNet-18 \citep{resnet2016CVPR}. Additionally, we train a ResNet-32 on CIFAR-100 as many CIL methods (e.g., \citep{ACIL2022NeurIPS} and \citep{RMM2021NeuriPS}) adopt this structure. All the experiments of CIL are conducted on  {an} NVIDIA GeForce RTX 3090 GPU with Python 3.9, PyTorch 1.13.1, and TorchVision 0.14.1 with PyCIL \citep{pycil}.

During the DS-BPT, we adopt the  {following} parameter settings. For the  {GBK acquisition stream}, we adopt the same training strategies for SSCL in \citep{SimSiam2021CVPR} and train the backbones for 1000, 600, and 100 epochs on CIFAR-100, ImageNet-100, and Imagenet-1k respectively, using SGD with a momentum of 0.9 and an initial learning rate of 0.1. The learning rates decay in a cosine  {manner}, and the value for weight decay is $5\times10^{-4}$. The batch size for training is 512. The projector and predictor  {comprise 2-layer MLPs with batch normalization incorporated after the FCNs}. During training, under the setting of stopping gradients in Simsiam \citep{SimSiam2021CVPR}, only the gradients of features produced by the network containing the predictor are backpropagated.  {For data augmentation in the GBK acquisition}, we use the combination of \textit{RandomResizedCrop}, \textit{RandomHorizontalFlip}, and \textit{ColorJitter}.  {For the SFD learning stream}, we follow DS-AL \citep{AAAI2024DSAL} to train the networks using SGD for 160 (90) epochs for ResNet-32 on CIFAR-100 (ResNet-18). The learning rate starts at 0.1 and is divided by 10 at epoch 80 (30) and 120 (60). We adopt a momentum of 0.9 and weight decay of $5\times10^{-4}$ ($1\times 10^{-4}$) with a batch size of 128.  {No special data augmentation exists in SFD learning}. 

For RED, we adopt the learning epochs $e=\{50,50,100\}$ and balancing factor $\lambda=\{0.5,0.5,0.9\}$ on CIFAR-100, ImageNet-100, and ImageNet-1k. For  {AL} in REAL and REAL-DS, we follow ACIL and DS-AL to adopt $\gamma=1e-5$ and $d_{\bm{B}}= 15k$.  {In REAL-DS, following \citep{AAAI2024DSAL}, the compensation ratio is adopted as $C = \{0.6,0.8,1.4\}$ on CIFAR100, ImageNet-100, and ImageNet-1k respectively.}

\textbf{Evaluation Protocol.} We adopt two metrics for evaluation, \textit{average incremental accuracy} and \textit{last-phase accuracy}. The overall performance is evaluated by the \textit{average incremental accuracy} (or average accuracy) $\mathcal{\bar A} = \frac{1}{K+1}{\sum}_{k=0}^{K}\mathcal{A}_{k}$, where $\mathcal{A}_{k}$ indicates the average test accuracy at phase $k$ obtained by testing on $\mathcal{D}_{0:k}^{\text{test}}$. A higher $\mathcal{\bar A}$ score is preferred at evaluations. {The other metric \textit{last-phase accuracy} $\mathcal{A}_{K}$ measures the last-phase performance of the network after completing all CIL tasks.} $\mathcal{A}_{K}$ is an important metric as it reveals the gap between CIL joint training, {a gap all CIL methods aim to close}.

\subsection{Comparison with State-of-the-Art Methods}

To validate the performance of our methods, {we compare them with both exemplar-based and exemplar-free methods}.  Results on CIFAR-100, ImageNet-100, and ImageNet-1k are {presented} in Table \ref{table:acc_cifar}, \ref{table:acc_I-100}, and \ref{table:acc_I-1k} respectively. 

\begin{table*}	
    \small
	\caption{Comparison of average accuracy $\mathcal{\bar A}$ and last-phase accuracy $\mathcal{A}_{K}$ among EFCIL and replay-based methods on CIFAR-100. We use both backbones of ResNet-32 and ResNet-18 for fair comparison. Data in Bold indicate the best within EFCIL methods and data \underline{underlined} are the best considering both EFCIL and EBCIL. ``EF" is the abbreviation of exemplar-free. ``$K$" indicates the number of phases.}
	\centering
		\begin{tabular}{clcc cccc cccc} 
			\toprule \\
			\multirow{2}{*}{Backbone}&\multirow{2}{*}{Method} &\multirow{2}{*}{EF?}& \multicolumn{4}{c}{$\mathcal{\bar A}$(\%)} &  & \multicolumn{4}{c}{$\mathcal{A}_{K}$(\%)}  \\ \cline{4-7} \cline{9-12} \\
            &&& K=5    & 10    & 25    & 50   &  & K=5      & 10     & 25     & 50     \\ \hline   
			\multirow{10}{*}{ResNet-32}
            &LUCIR \citep{LUCIR2019_CVPR} &$\times$ & 63.17 & 60.14 & 57.54 & 55.57 &&54.30&50.30&48.35&42.10 \\
            &PODNet \citep{podnet2020ECCV} &$\times$ & 64.83 & 63.19  & 60.72 & 59.10 &&54.60&53.00&51.40&46.64 \\
            &POD+AANets \citep{AANet_2021_CVPR} &$\times$ & 66.31 & 64.31 &62.31 & 59.15 &&59.39&57.37&53.55&49.40 \\
            &POD+AANets+RMM \citep{RMM2021NeuriPS} &$\times$ & 68.36 & 66.67& 64.12 & 63.54 &&59.00&59.03&56.50&54.3 \\
            &FOSTER \citep{FOSTER2022ECCV} &$\times$ & \underline{72.58}  & {67.95} & 65.67 & 59.57 &&\underline{65.57} &58.44&56.15&52.23 \\ 
            
            &LwF \citep{LwF2018TPAMI} &{\color{red}$\checkmark$ }& 49.59 & 48.47 & 45.75 & 10.19 &&40.40&40.19&38.25&2.09 \\
            &ACIL \citep{ACIL2022NeurIPS}&{\color{red}$\checkmark$} & 66.30 & 66.07 & 65.95 & 66.01 &&57.78&57.79&57.65&57.83 \\      
            &DS-AL \citep{AAAI2024DSAL} &{\color{red}$\checkmark$} & 66.46 & 66.22 & 66.13 & 66.33 &&58.14&58.12&58.19&58.37 \\   \cline{2-12} 
			&  \textbf{REAL (ours)}&{\color{red}$\checkmark$} & 68.78 &68.62 &68.60 &68.41 &&61.19 &	61.15 &	61.46 &	61.32 \\ 
			&  \textbf{REAL-DS (ours)}&{\color{red}$\checkmark$}&\textbf{68.88}&{\underline{\textbf{68.75}}}&\underline{\textbf{68.79}}&\underline{\textbf{68.42}}&&\textbf{61.62}&\underline{\textbf{61.76}}&\underline{\textbf{62.18}}&\underline{\textbf{61.87}} \\ 
            \hline
			\multirow{7}{*}{ResNet-18}
			&PASS \citep{PASS2021CVPR} &{\color{red}$\checkmark$}& 63.88 & 60.07 & 56.86 & 41.11&&55.75&49.13&44.76&28.02 \\
            &PRAKA \citep{PRAKA2023ICCV} &{\color{red}$\checkmark$} & 68.59 & 68.51 & 62.98 & 43.09 &&61.52&60.12&51.15&27.14 \\ 
            &ACIL \citep{ACIL2022NeurIPS}&{\color{red}$\checkmark$} & 68.07 & 68.08& 68.07 & 67.94 && 60.65& 61.03& 61.41 &61.09 \\
            &DS-AL \citep{AAAI2024DSAL} &{\color{red}$\checkmark$} & 68.55 & 68.23 & 68.20 & 68.17 && 61.35&61.31&61.41&61.41 \\ \cline{2-12}
            &  \textbf{REAL (ours)}&{\color{red}$\checkmark$}& {\textbf{72.35}}&{\underline{\textbf{72.41}} }&{\underline{\textbf{72.28}}}&{\underline{\textbf{72.25}} }&&65.73&65.86&65.81&65.76 \\ 
			&  \textbf{REAL-DS (ours)} &{\color{red}$\checkmark$}&{72.24}&{72.35}&{72.20}&{72.23}&&\underline{\textbf{66.27}}&\underline{\textbf{66.74}}&\underline{\textbf{66.57}}&\underline{\textbf{66.35}}\\
			\bottomrule
		\end{tabular} 
 \label{table:acc_cifar}
	\end{table*}

\textbf{Baselines.} For exemplar-free methods, we validate REAL with LwF \citep{LwF2018TPAMI}, PASS \citep{PASS2021CVPR},  PRAKA \citep{PRAKA2023ICCV}, ACIL \citep{ACIL2022NeurIPS} and DS-AL \citep{AAAI2024DSAL}. Among these techniques, LwF is a regularization-based method {that uses KD to construct a regularization term} in objective function, PASS and PRAKA are within the category of prototype-based methods, and ACIL and DS-AL are ACL techniques.
{Additionally, we include} exemplar-based methods, i.e., LUCIR \citep{LUCIR2019_CVPR}, PODNet \citep{podnet2020ECCV}, AANets \citep{AANet_2021_CVPR}, RMM \citep{RMM2021NeuriPS} and FOSTER \citep{FOSTER2022ECCV}, as counterpart. For the compared methods, we follow the official implementations from their papers.
 
\textbf{Validation on CIFAR-100.} As depicted in Table \ref{table:acc_cifar}, on CIFAR-100, we validate REAL with ResNet-18 and ResNet-32 {as several compared methods are implemented with different backbones}.  The results for CIFAR-100 in Table \ref{table:acc_cifar} underscore the competitive performance of EBCIL methods, owing to the benefits of sample storage and exemplar replay. Notably, with a smaller $K$ (e.g., $K=5, 10$), FOSTER achieves the best results among all compared methods. However, the ACL techniques (ACIL and DS-AL) marginally lag behind ``POD+AANets+RMM" in these settings. Through representation enhancement, our REAL elevates ACL methods to approach or even surpass the {performance of} state-of-the-art EBCIL methods with a small $K$ of 5. For instance, REAL-DS demonstrates superior results in both average accuracy and final phase accuracy when compared {with} ``POD+AANets+RMM". With a larger $K$ of 10 or more, REAL-DS even outperforms the leading exemplar-based method, FOSTER, and maintains this trend with increasing $K$. Among the EFCIL methods, the compared ACL methods (i.e., ACIL and DS-AL) have already shown competitive performance on various settings of $K$ {with both} ResNet-18 and ResNet-32. As a new technique within the ACL family, our REAL can have improved performance and achieve best performance among the EFCIL methods. {For example, the average accuracy results of REAL-DS with $K=5,10,25$, and $50$ with ResNet-18, are 72.24\%, 72.35\%, 72.20\%, and 72.23\% respectively, surpassing the previous ACL baseline DS-AL with 3.69\%, 4.12\%, 4.00\%, and 4.06\% respectively}. {Notably}, as $K$ increases, all EBCIL and EFCIL methods except ACL experience a {notable performance decline}. For instance, the final phase accuracy results of FOSTER and PRAKA drop from 68.57\% and 61.52\% to 52.23\% each when $K$ rises from 5 to 50. {By} contrast, the ACL techniques, including REAL and REAL-DS, maintain consistent performance irrespective of the number of learning phases. This observation showcases that REAL maintains the invariance of phases inherent in ACL and take advantage of this property in large learning phases.

\textbf{Validation on ImageNet-100 and ImageNet-1k.} As {indicated} in Table \ref{table:acc_I-100}, the leading patterns of REAL and REAL-DS persist when compared {with} EFCIL methods. Notably, REAL outperforms the current state-of-the-art EFCIL method DS-AL by significant margins in both average incremental accuracy and final phase accuracy. The advantage of maintaining weight invariance across phases {enables} REAL and REAL-DS to deliver exceptional performance even with {larger $K$ values}. When compared with EBCIL methods, the trend of leading in scenarios with large $K$ is consistent with that observed on CIFAR-100. EBCIL methods excel in leveraging exemplar replay in settings with small $K$, while REAL {leads when $K$ reaches 10 or higher}. On ImageNet-1k, REAL and REAL-DS continue to outperform other EFCIL methods, and the leading pattern compared to EBCIL methods aligns with observations on ImageNet-100.

\begin{table*}	
    \small
	\caption{Comparison of average accuracy $\mathcal{\bar A}$ and last-phase accuracy $\mathcal{A}_{K}$ on ImageNet-100. Data in Bold indicate the best within EFCIL methods and data \underline{underlined} are the best considering both EFCIL and EBCIL. ``EF" is the abbreviation of exemplar-free. ``$K$" indicates the number of phases.}
	\centering
		\begin{tabular}{clcc cccc cccc}
			\toprule \\
			\multirow{2}{*}{Backbone}&\multirow{2}{*}{Method} &\multirow{2}{*}{EF?}& \multicolumn{4}{c}{$\mathcal{\bar A}$(\%)} &  & \multicolumn{4}{c}{$\mathcal{A}_{K}$(\%)}  \\ \cline{4-7} \cline{9-12} \\
            &&& K=5    & 10    & 25    & 50   &  & K=5      & 10     & 25     & 50     \\ \hline   
			\multirow{13}{*}{ResNet-18}
            &LUCIR \citep{LUCIR2019_CVPR}&$\times$ & 71.32 & 68.57 & 61.57 & 51.08 &&60.00&57.10&49.26&43.13 \\
            &PODNet \citep{podnet2020ECCV}&$\times$ & 75.90 &73.70&66.58&59.36&&67.60 &65.00&54.30&44.40  \\
            &POD+AANets \citep{AANet_2021_CVPR}&$\times$ & 76.96 & 75.58 & 71.78 & 64.63 &&69.40&67.47&63.69& 51.00 \\
            &POD+AANets+RMM \citep{RMM2021NeuriPS} &$\times$ & 79.52 & \underline{78.47} & 76.54 & 71.21 &&73.80&\underline{71.40}&68.84& 56.50 \\ 
            &FOSTER \citep{FOSTER2022ECCV}&$\times$ & \underline{79.94} & 77.73 &  71.79 & 66.77 && \underline{74.52}&70.54& 62.70 &59.20 \\  
            &LwF \citep{LwF2018TPAMI} &{\color{red}$\checkmark$ }& 54.12 & 47.90 & 44.45 & 13.13 &&40.10&36.10&34.12&2.30 \\
            &PASS \citep{PASS2021CVPR}&{\color{red}$\checkmark$}& 72.24 & 67.97 & 52.02 & 30.59 &&61.76&57.38&37.46&18.22 \\
            &PRAKA \citep{PRAKA2023ICCV}&{\color{red}$\checkmark$} & 63.66 & 62.75 & 58.86 & 41.24 &&55.77&54.77&46.49&21.80 \\ 
            
            &ACIL \citep{ACIL2022NeurIPS}&{\color{red}$\checkmark$} & 74.81&74.76 &74.59&74.57&&66.98&67.42&67.16&67.38  \\   
            &DS-AL \citep{AAAI2024DSAL}&{\color{red}$\checkmark$} & 75.08 & 75.89 & 74.70 & 74.78 &&67.54&67.38&67.32&67.50 \\ \cline{2-12}
			&\textbf{REAL (ours)}&{\color{red}$\checkmark$}& 76.48&	76.39 &76.14 &76.20 &&68.94 &69.18 &69.00&69.02 \\ 
			&\textbf{REAL-DS (ours)} &{\color{red}$\checkmark$}& \textbf{77.30}&\textbf{77.32}&\underline{\textbf{77.19}}&\underline{\textbf{77.11}}&&\textbf{69.52}&\textbf{69.82}&\underline{\textbf{69.76}}&\underline{\textbf{69.46}} \\
			\bottomrule                                                                                   
		\end{tabular} 
 \label{table:acc_I-100}
\end{table*}

\begin{table*}
    \small
	\caption{Comparison of average accuracy $\mathcal{\bar A}$ and last-phase accuracy $\mathcal{A}_{K}$ on ImageNet-1k. Data in Bold indicate the best within EFCIL methods and data \underline{underlined} are the best considering both EFCIL and EBCIL. EF is the abbreviation of exemplar-free. ``$K$" indicates the number of phases. 
    }
	\centering
		\begin{tabular}{clcc ccc ccc}
			\toprule \\
			\multirow{2}{*}{Backbone} & \multirow{2}{*}{Method} &\multirow{2}{*}{EF?}& \multicolumn{3}{c}{$\mathcal{\bar A}$(\%)} &  & \multicolumn{3}{c}{$\mathcal{A}_{K}$(\%)}  \\ \cline{4-7} \cline{8-10} \\
            &&& K=5    & 10    & 25    &  & K=5      & 10     & 25         \\ \hline   
            \multirow{13}{*}{ResNet-18} 
            &LUCIR \citep{LUCIR2019_CVPR}&$\times$ & 64.45 & 61.57 & 56.56  &&56.60&51.70&46.23\\
            &PODNet \citep{podnet2020ECCV}&$\times$ & 66.43 & 63.20 & 59.17  &&58.90&55.70&50.51 \\
            &POD+AANets \citep{AANet_2021_CVPR}&$\times$ & 67.73 & 64.85 & 61.78  &&60.84&57.22&53.21 \\
            &POD+AANets+RMM \citep{RMM2021NeuriPS} &$\times$ & \underline{69.21} & 67.45 & 63.93  &&\underline{62.50}&\underline{60.10}&55.50 \\
            &FOSTER \citep{FOSTER2022ECCV}&$\times$ & 61.65 & 60.74 & 57.64   && 51.45& 49.60& 46.80  \\ 
            &LwF \citep{LwF2018TPAMI}&{\color{red}$\checkmark$ }& 44.35 & 38.90 & 36.87 &&34.20&30.10&30.01 \\ 
            &PASS \citep{PASS2021CVPR}&{\color{red}$\checkmark$}& 58.44 &56.44&52.41 &&49.83 &47.71&42.55  \\
            &PRAKA \citep{PRAKA2023ICCV}&{\color{red}$\checkmark$} & 51.71 &49.48 &45.56 &&44.35 &41.85 &36.54  \\ 
            &ACIL \citep{ACIL2022NeurIPS}&{\color{red}$\checkmark$} & 65.34 & 64.84 & 61.78 &&56.11&55.43&55.43 \\
            &DS-AL \citep{AAAI2024DSAL}&{\color{red}$\checkmark$} & 67.15 & 66.98 & 66.88 &&58.12&58.20&58.12 \\  \cline{2-10}
			&\textbf{REAL (ours)}&{\color{red}$\checkmark$}& 67.41& 67.18& 67.15&&58.06&58.06&58.14\\ 
			&\textbf{REAL-DS (ours)} &{\color{red}$\checkmark$}&\textbf{68.04}&\underline{\textbf{67.78}}&\underline{\textbf{67.60}}&&\textbf{59.39}&\textbf{59.35}&\underline{\textbf{59.33}} \\
			\bottomrule
		\end{tabular}
     \label{table:acc_I-1k}
\end{table*}

\begin{figure*}
	\centering
	\includegraphics[width=1\linewidth]{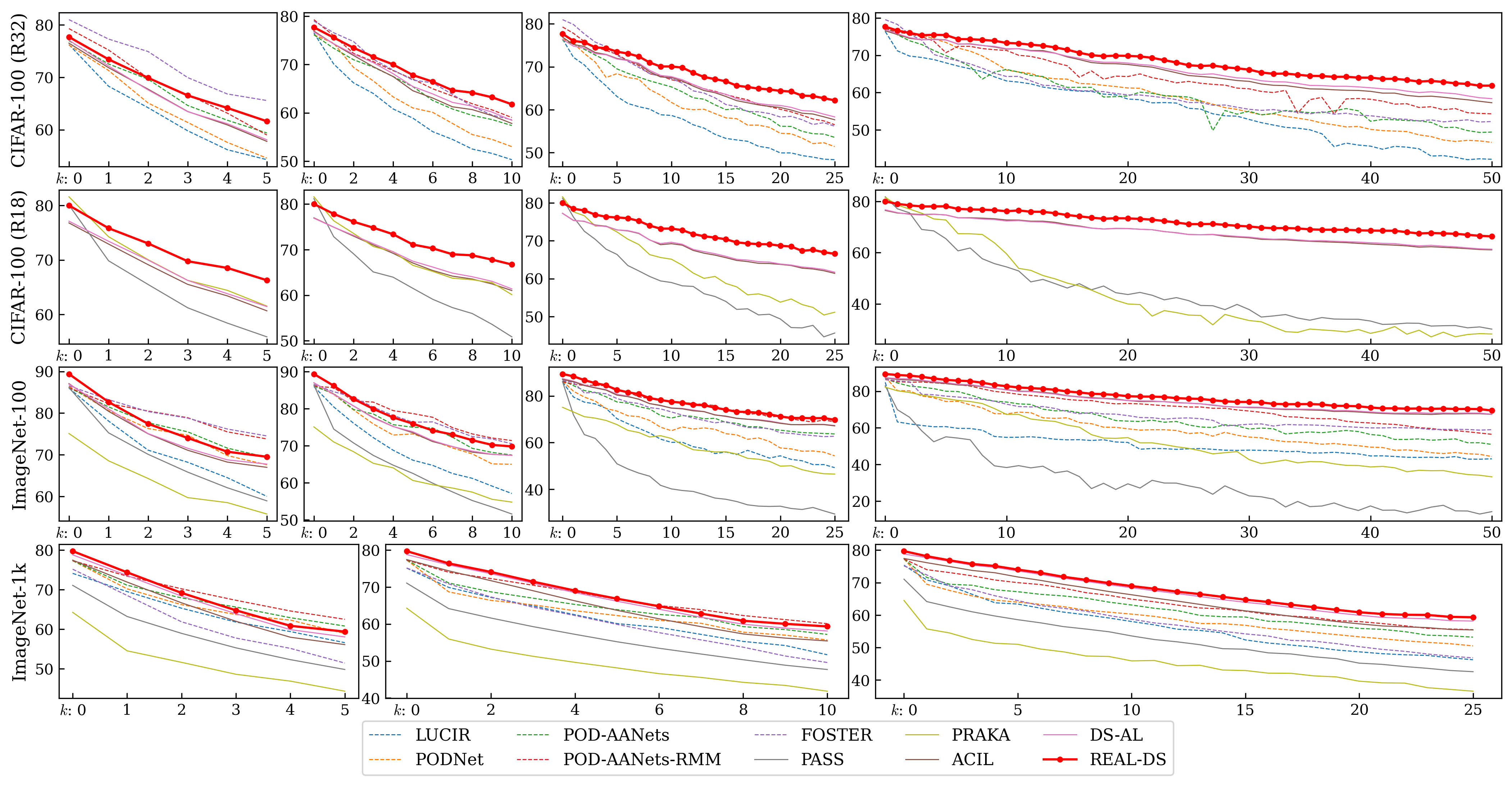}
	\caption{Phase-wise accuracy of compared methods on benchmark datasets with $K = 5, 10, 25$, and $50$ on CIFAR-100 and ImageNet-100. The results of ImageNet-1k with $K = 5, 10, 25$ are also included. The dash lines are the curves for exemplar-based methods. ``CIFAR-100 (R32)" denotes the results on CIFAR-100 with backbone of ResNet-32, while ``R18" indicates using the ResNet-18. Experiments without specific denotation use ResNet-18 as the backbone. }
	\label{fig:phasewiseacc}
\end{figure*}

\textbf{Evolution Curves of Phase-wise Accuracy.} {We present the evolution curves of phase-wise accuracy on all datasets in Figure \ref{fig:phasewiseacc} to comprehensively vasualize the results}. We plot the results phase-wise accuracy of LUCIR, PODNet, POD+AANets, POD+AANets+RMM, FOSTER, PASS, PRAKA, ACIL, DS-AL and our proposed REAL-DS. The EBCIL methods are plotted in dash lines. As illustrated in Figure \ref{fig:phasewiseacc}, REAL-DS achieves the highest performance among all benchmark datasets when compared with EFCIL methods. The leading trend commences from the base phase and is consistently maintained through to the final phase, {underscoring exceptional performance of REAL-DS }during the incremental learning process. When contrasted with EBCIL methods, REAL-DS exhibits a more gradual decline trend when a larger $K=25,50$ is employed. {This trend indicates that REAL-DS effectively addresses larger learning phases and surpass the compared methods}. 

{
\subsection{Forgetting Analysis}\label{subsection:forgetting analysis}
We propose a comparative study on the average forgetting rate $\bar{\mathcal{F}}$ \citep{FOAL2024NeurIPS} between REAL and the existing methods to analyze REAL's performance of reducing forgetting. A lower average forgetting rate shows a better performance of resisting forgetting during the incremental learning process. The average forgetting rate is defined as $\mathcal{\bar F} = \frac{1}{K}{\sum}_{k=1}^{K}\mathcal{F}_{k}$, where $\mathcal{F}_{k}$ {denotes} the forgetting rate at phase $k$ obtained by testing the model of phase $k$ on previous data then comparing this accuracy with the optimal previous results. The $\mathcal{F}_{k}$ can be obtained by:
\[
\begin{array}{c}
\mathcal{F}_{k}=\frac{1}{k} \sum_{j=0}^{k-1} {f}_{k, j}, \quad
{f}_{k, j}=\max _{l \in\{1, \ldots, k-1\}}\left(\mathcal{A}_{l, j}\right)-\mathcal{A}_{k, j}, \quad \forall j<k ,
\end{array}
\]
where $\mathcal{A}_{k, j}$ denotes accuracy evaluated on the test set of phase $j$ after training the network at phase $k$.

As {indicated} in Table \ref{table:forgetting}, REAL and REAL-DS demonstrate superior performance of reducing forgetting when compared with other methods on both CIFAR-100 and ImageNet-100. For example, REAL achieves the $\bar{\mathcal{F}}$ of 4.66\%, 4.05\%, 3.96\%, and 4.22\% on CIFAR-100 with K=5, 10, 25, and 50 respectively, outperforming PASS and even the exemplar-based methods (PODNet and FOSTER) with large gaps. When compared with ACL techniques, REAL and REAL-DS have lower forgetting rates, demonstrating the effect of well mitigating forgetting introduced by RED and FFD.  

\begin{table}	
    \footnotesize
	\caption{{Comparison of average forgetting $\mathcal{\bar F}(\%)$ on CIFAR-100 and ImageNet-1k. Data in Bold indicate the best results within compared methods. ``$K$" indicates the number of phases. }
    }
	\centering
    \resizebox{0.48\textwidth}{!}{
    {
		\begin{tabular}{lc cccc cccc}
			\toprule \\
			 \multirow{2}{*}{Method} & \multicolumn{4}{c}{CIFAR-100} &  & \multicolumn{4}{c}{ImageNet-100}  \\ \cline{2-5} \cline{7-10} \\
            & K=5    & 10    & 25    & 50 &  & K=5      & 10     & 25   & 50     \\ \hline   
            PODNet \citep{podnet2020ECCV}&15.64 & 20.71 & 24.93 & 30.07  &&14.91&18.26&24.35&26.29 \\
            FOSTER \citep{FOSTER2022ECCV}&13.12 & 18.28 & 21.58 & 27.45   && 12.14 & 15.21& 25.22 & 35.39  \\ 
            PASS \citep{PASS2021CVPR}&16.09& 20.05 &23.18&30.71 &&10.44 &18.43&28.32 & 35.92  \\
            ACIL \citep{ACIL2022NeurIPS}&5.46 & 4.61 & 4.85 & 5.41 &&6.54&7.09&9.09& 9.52 \\
            DS-AL \citep{AAAI2024DSAL}&6.41 &5.98 & 5.80 & 6.08 &&8.07&9.79&11.34&11.69 \\  \cline{1-10}
			\textbf{REAL (ours)}& \textbf{4.66}& \textbf{4.05}& \textbf{3.96}&\textbf{4.22}&&\textbf{5.68}&\textbf{7.79}&\textbf{7.13}&\textbf{8.49}\\ 
			\textbf{REAL-DS (ours)} &5.81&4.77&4.55& 4.88 &&6.16&8.20&8.71&10.31 \\
			\bottomrule
		\end{tabular}
    }
    }
     \label{table:forgetting}
\end{table}

}

\subsection{Ablation Study}\label{subsection:5.4}

\textbf{Ablation Study on RED and FFB.} In this section, we explore the efficacy of each component in the proposed REAL model, specifically the RED and FFB modules. We perform the experiments on CIFAR-100 and ImageNet-100 using ResNet-18 in the 5-phase CIL setting and present the results in Table \ref{table_ablation}. To validate the effectiveness of RED, we utilize the backbones with GBK and SFD as two distinct knowledge sources and merge two streams with RED. As {indicated} in the Table \ref{table_ablation}, without FFB, the backbone with SFD outperforms that with GBK on CIFAR-100 in terms of both $\mathcal{A_K}$ and $\mathcal{\bar A}$. {Contrastingly}, the results are reversed on ImageNet-100, highlighting the advantages of general knowledge on larger datasets. {This observation indicates that general knowledge can be more effectively on a larger dataset owing to the abundance of training samples}, while a smaller dataset relies more on labels to guide the learning process. By utilizing RED to merge these two types of knowledge, the model achieves significantly better performance {than} models trained on single knowledge sources. This underscores that the knowledge fusion enabled by RED leverages the strengths of both knowledge types, enhancing {the  performance of base model in incremental learning}. {Since SFD is obtained by the cross entropy loss and the GBK is obtained by training with SSCL loss, this study demonstrates the benefit of leveraging both types of training techniques.} Furthermore, upon incorporating FFB, performance with all backbones is improved, indicating that FFB {can effectively generate} informative features.

\begin{table}[!h]
    \footnotesize
   \caption{Ablation study of the knowledge components in REAL.}
    \begin{tabular}{lccccc}
				\hline
				 \multirow{2}{*}{\textbf{Knowledge Sources}} &\multirow{2}{*}{\textbf{FFB}}&\multicolumn{2}{c}{\textbf{CIFAR-100}} & \multicolumn{2}{c}{\textbf{ImageNet-100}} \\ \cline{3-6}
                 & & $\mathcal{A_K}$ (\%)  & $\mathcal{\bar A}$ (\%) & $\mathcal{A_K}$ (\%)  & $\mathcal{\bar A}$ (\%) \\ 
				\hline 
				\multirow{2}{*}{\textbf{GBK}} &$\times$ & 57.01 & 62.66& 67.42& 74.48   \\ 
                & {\color{red}$\checkmark$}&60.22 & 64.88& 67.98& 74.76\\ \hline
				\multirow{2}{*}{\textbf{SFD}} &$\times$ & 60.91& 68.21& 65.64& 73.56  \\ 
                & {\color{red}$\checkmark$}& 62.41& 69.23& 67.28& 74.36\\ \hline
                \multirow{2}{*}{\textbf{Merging with RED}} &$\times$ & 64.98& 71.85& 68.36& 76.01  \\ 
                & {\color{red}$\checkmark$}& \textbf{65.72}& \textbf{72.34}& \textbf{68.94}& \textbf{76.48}\\ \hline
		\end{tabular} 
		\centering
		\label{table_ablation}
\end{table}	

{
\textbf{RED Surpasses Directly Using Two Base Training Losses.} In the base training, REAL captures the general knowledge and class-specific knowledge via SSCL and SL respectively (DS-BPT), and then merges them via RED. A simpler approach is to directly utilize the SSCL and SL loss jointly to formulate a joint training loss during base training. Here, we use the same parameters and conduct a comparative study on RED and the training with joint training loss on CIFAR-100. The joint loss can be formulated as $\mathcal{L}_{\text{joint}} = \alpha \mathcal{L}_{\text{label}} + \beta \mathcal{L}_{\text{GBK}}$, where $\alpha$ and $\beta$ are the balancing coefficients. As indicated in Table \ref{table_two_loss}, REAL outperforms the results of various combinations of $\alpha$ and $\beta$. This observation showcases the superiority of RED in merging the general knowledge and class-specific knowledge. The reason that joint training loss falls behind may be that, these two goals may have conflicts and this training can lose the soft knowledge of SFD. Since the SSCL considers the sample-wise similarities while the SL loss primarily focuses on the category-wise discrepancy, they may affect each other and satisfying these two objectives together in the early stages may be hard to achieve. Also, RED leverages the soft knowledge of SFD, leading to a superior performance when compared with directly using the joint objective.  
}
\begin{table}[!h]
    \footnotesize
    \caption{{Comparative study of RED and directly using two base training losses of SSCL and SL.}}
    \begin{tabular}{lcc}  
				\toprule
                 \textbf{Base Training Loss} & $\mathcal{A_K}$ (\%)  & $\mathcal{\bar A}$ (\%)  \\ 
				\hline 
				$\mathbf{\alpha=1.0, \beta=0.1}$ &53.40 & 64.01   \\ 
                $\mathbf{\alpha=1.0, \beta=0.5}$ &54.89 & 65.35   \\ 
                $\mathbf{\alpha=1.0, \beta=1.0}$ &56.95 & 66.93   \\ 
                $\mathbf{\alpha=0.5, \beta=1.0}$ &56.51 & 66.74   \\ 
                $\mathbf{\alpha=0.1, \beta=1.0}$ &49.27 & 56.74   \\ \hline        
                \textbf{REAL (ours)}& \textbf{65.73}& \textbf{72.35}\\ \bottomrule
		\end{tabular} 
		\centering
	       \label{table_two_loss}
\end{table}	

{
\textbf{Comparative Study between REAL and SupCon.}  Supervised contrastive loss (SupCon) \citep{SupCon2020NeurIPS} represents a loss of contrastive learning with the label information. Here, we propose a comparative study between REAL and SupCon on CIFAR-100 for $K=5$. The results of various settings of epoch (e) and batch size (bsz) are tabulated in Table \ref{table:supcon}. SupCon achieves similar results to the GBK only (e.g., 56.80\% and 64.31\% of SupCon($e=200, bsz=2048$) vs 57.01\% and 62.66\% of $\mathcal{A_K}$ and $\bar{\mathcal{A}}$ respectively) but falls behind REAL, which demonstrates that, though SupCon is a contrastive learning method based on the label information, it may not benefit as effective as the CE loss.  REAL outperforms the results obtained by different parameter settings and demonstrates its effectiveness of merging general knowledge stems from the contrastive learning and the class-specific knowledge from SL.     
}

\begin{table}[!h]
    \footnotesize
   \caption{Comparative study of REAL and  using SupCon loss in the base training. ``e" denotes the epochs and ``bsz" denotes the batch size.}
    
    \begin{tabular}{lcc}  
				\toprule
                 \textbf{Base Training Loss} & $\mathcal{A_K}$ (\%)  & $\mathcal{\bar A}$ (\%)  \\ 
				\hline 
				\textbf{SupCon($e=200, bsz=512$)} &54.52 & 63.33   \\ 
                \textbf{SupCon($e=200, bsz=1024$)} &54.68 & 63.65   \\
                \textbf{SupCon($e=200, bsz=2048$)} &56.80 & 64.31   \\
                \textbf{SupCon($e=500, bsz=512$)} &56.49 & 65.96  \\
                \textbf{SupCon($e=500, bsz=1024$)} &55.31 & 64.55 \\
                \textbf{SupCon($e=500, bsz=2048$)} &54.77 & 63.94   \\ \hline    
                \textbf{GBK only} & 57.01 & 62.66 \\
                \textbf{REAL (ours)}& \textbf{65.73}& \textbf{72.35}\\ \bottomrule
		\end{tabular} 
		\centering
	\label{table:supcon}
\end{table}	

\begin{figure}
	\centering
	\includegraphics[width=1.0\linewidth]{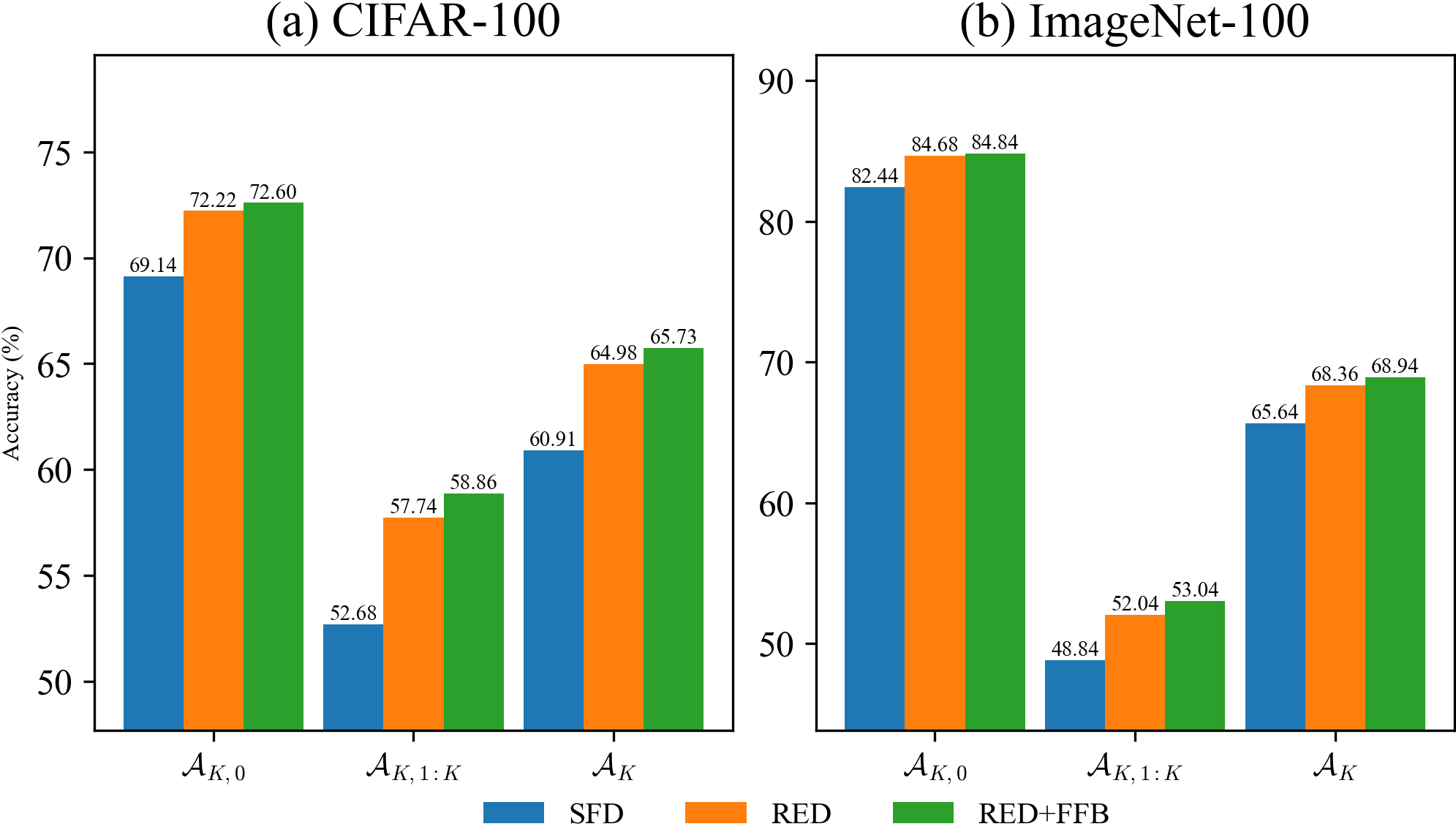}
	\caption{Analysis on stability and ;plasticity by ablating RED and FFB.}
	\label{fig:FHLH}
\end{figure}

\textbf{REAL Improves Both Stability and Plasticity.} Here, we investigate the impact of each component of REAL on the stability and plasticity by ablating RED and FFB. To demonstrate this, after training on all phases, we evaluate the models on the base dataset $\mathcal{D}_{0}^{\text{test}}$ (i.e., the first half) to show the stability and the CIL dataset $\mathcal{D}_{1:K}^{\text{test}}$ (i.e., the other half) to evaluate the plasticity. The experiments are conducted on CIFAR-100 and ImageNet-100. We report the accuracy on base dataset $\mathcal{A}_{K,0}$, the accuracy on CIL dataset $\mathcal{A}_{K,1:K}$, and last phase accuracy $\mathcal{A}_{K}$ in Figure \ref{fig:FHLH}. Without any modification, the baseline of only SFD obtains relatively low performance on both $\mathcal{A}_{K,0}$ and $\mathcal{A}_{K,1:K}$ and then results in a low performance of $\mathcal{A}_{K}$. When {RED is included, the results of both $\mathcal{A}_{K,0}$ and $\mathcal{A}_{K,1:K}$ improve significantly}, demonstrating RED can enhance both stability and plasticity. The enhanced plasticity can be attributed to the discriminative capabilities provided by GBK acquisition. {In particular}, GBK aids in distinguishing data categories not encountered by the backbone during the base phase, thereby facilitating classification during incremental learning. On the stability front, REAL enriches data representations during the base phase, thereby reinforcing foundational knowledge. {With the inclusion of FFB}, the performance can be further improved in terms of both stability and plasticity. FFB fuses both the transferable and specific features, {ultimately improving the overall stability and plasticity of the model}.


\begin{figure}[!h]
	\centering
	\includegraphics[width=1.0\linewidth]{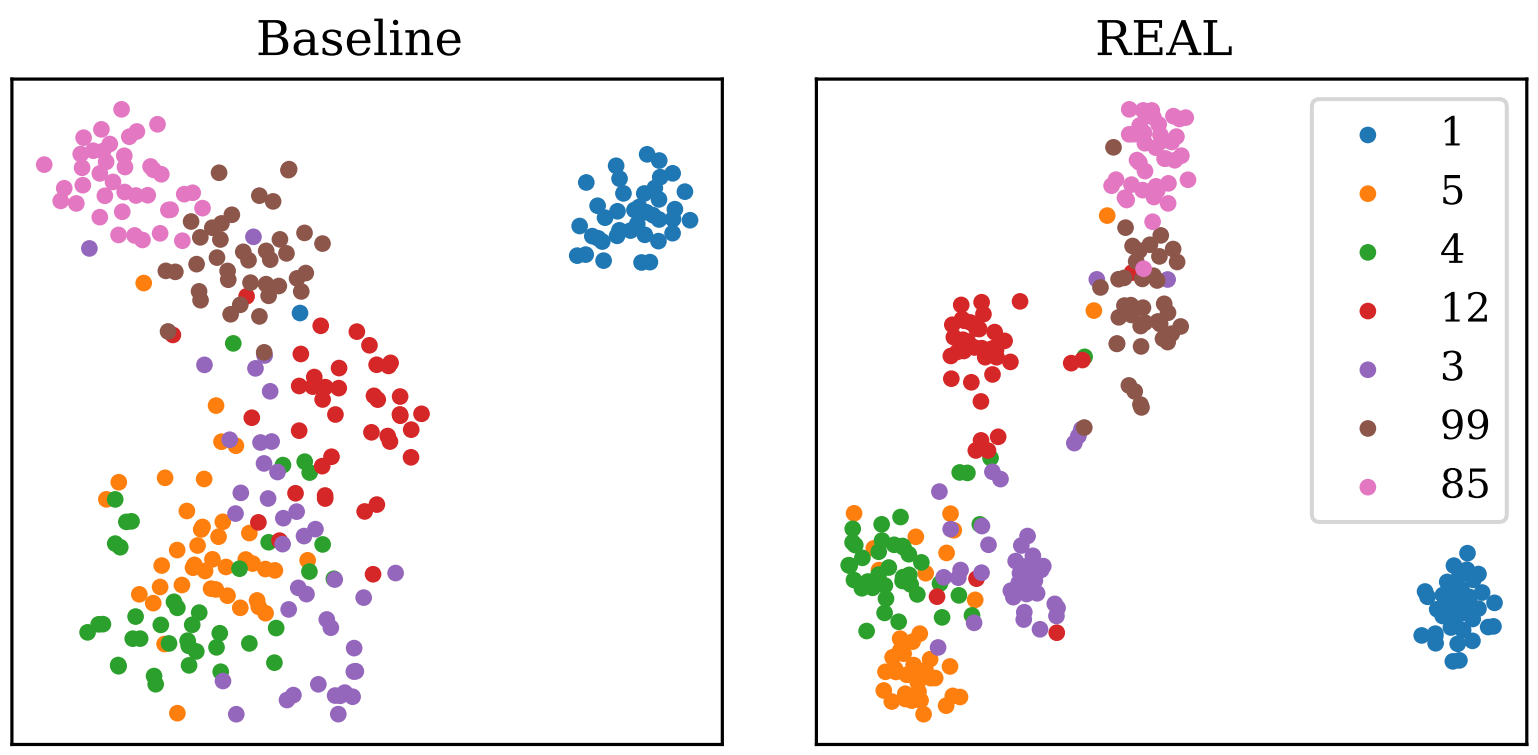}
	\caption{T-SNE plot of feature vectors extracted on ImageNet-100. The features extracted with REAL within a class are more compact compared with those in the baseline.}
	\label{fig:tsne_red}
\end{figure}

\textbf{Visualization of Representation Enhancement.} To demonstrate the effect of RED and FFB, we conduct a qualitative study by visualizing feature vectors in 2D with t-SNE \citep{tSNE2008JMLR}. {The feature vectors are extracted on the validation set of ImageNet-100 by randomly choosing 5 classes from the base dataset and 2 classes from CIL dataset}. As depicted in Figure \ref{fig:tsne_red}, with REAL, the feature vectors within a category exhibit a clustering effect, where vectors belonging to the same category are pulled closer together compared to the baseline model without RED and FFB. {This clustering phenomenon is evident within base classes (e.g., classes 3 and 12) and extends to classes that are not directly learned in the base phase (i.e., classes 85 and 99)}. By bringing features within the same category closer together, REAL facilitates the construction of more discriminative decision boundaries, thereby enhancing the overall performance.

{
\textbf{Plug and Play Technique for ACL Branch.} To validate the compatibility of REAL with more ACL methods, we conduct further experiments of incorporating REAL into another ACL method GKEAL \citep{GKEAL2023CVPR} on CIFAR-100 and ImageNet-100. As indicated in Table \ref{table:gkeal}, when incorporating REAL into GKEAL, the results on both the CIFAR-100 and ImageNet-100 are improved across all $K$ values. For example, on CIFAR-100, REAL can improve the $\mathcal{A}_{K}$ GKEAL from 57.65\% to 61.54\%. This observation further demonstrates that REAL can be a beneficial plug-and-play technique for ACL methods. 
}

\begin{table}[h]
    \footnotesize
	\caption{{Comparison of the average accuracy and  $\mathcal{\bar A}(\%)$ and the last-phase accuracy $\mathcal{A}_{K}$ between GKEAL and introducing REAL to GKEAL (denoted as REAL-GKEAL) on CIFAR-100 and ImageNet-1k.``$K$" indicates the number of phases.}
    }
	\centering
    \resizebox{0.48\textwidth}{!}{
    {
		\begin{tabular}{cl ccc c ccc}
			\toprule 
            \multirow{2}{*}{Dataset} & \multirow{2}{*}{Method}  & \multicolumn{3}{c}{$\mathcal{A}_K$} &  & \multicolumn{3}{c}{$\bar{\mathcal{A}}$} \\ \cline{3-5} \cline{7-9}
			  & & K=5    & 10    & 25   && K=5    & 10    & 25   \\ \hline   
            \multirow{2}{*}{CIFAR-100} &
			GKEAL& 57.65& 57.65& 57.65 && 66.72 & 66.54 & 66.43\\ 
			&\textbf{REAL-GKEAL} &\textbf{61.54}&\textbf{61.54}&\textbf{61.54} &&\textbf{70.82} &\textbf{70.60}&\textbf{70.41} \\ \hline   
            \multirow{2}{*}{ImageNet-100} & 
			GKEAL& 64.52&  64.52&  64.52 && 73.06 & 72.85 & 72.77\\ 
			&\textbf{REAL-GKEAL} &\textbf{64.96}&\textbf{64.96}&\textbf{64.96} && \textbf{74.87} & \textbf{74.62} & \textbf{74.50} \\
			\bottomrule
		\end{tabular}
    }
    }
     \label{table:gkeal}
\end{table}

\subsection{Hyperparameter Analysis} \label{sec:hyper}
\begin{figure}[h!]
	\centering
	\includegraphics[width=1.0\linewidth]{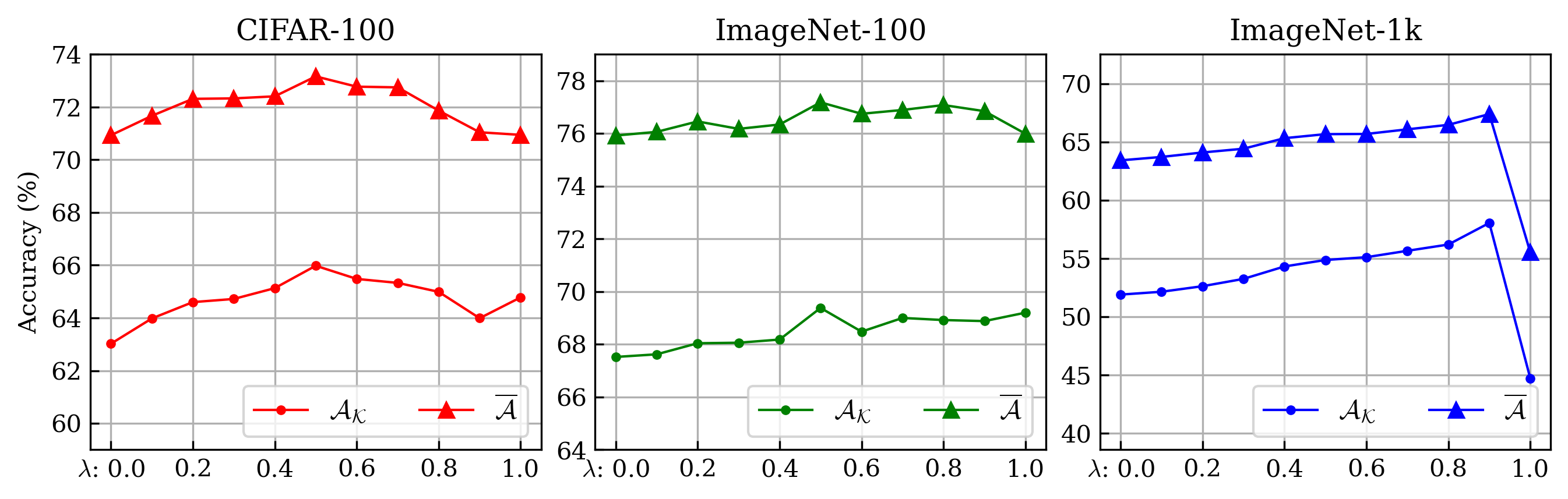}
	\caption{Accuracy on benchmark datasets with varying $\lambda$.}
	\label{fig:acc_kd}
\end{figure}

\textbf{Balancing Labels and Teacher Model with $\lambda$.} To analyze the impact of $\lambda$, we conduct the experiments with a 5-phase CIL setting {and the results of $\mathcal{A}_{K}$ and $\mathcal{\bar A}$ from $\lambda = 0.0$ to $\lambda = 1.0$ are plotted in Figure \ref{fig:acc_kd}}. Across all three benchmark datasets,  the performance initially improves with larger $\lambda$ values, reaching a peak before decreasing. This trend highlights the balance between knowledge derived from the labels and the feature distribution learned by the backbone. {The optimal $\lambda$ values that result in the best performance increase as the dataset scales up}, transitioning from $\lambda=0.5$ on CIFAR-100 and ImageNet-100 to $\lambda=0.9$ on ImageNet-1k. This pattern shows that models trained on larger datasets tend to prefer knowledge from the feature distribution learned in the backbone. {Feature distributions acquired from extensive datasets, for example, the ImageNet-1k, are potent and highly transferable owing to the abundance of training samples available}. During the RED process, the SFD knowledge is transferred to the backbone with GBK and contributes more than the labels solely.  {Notably, utilizing only labels ($\lambda=0.0$) or solely relying on the backbone with SFD ($\lambda=1.0$) results in less effective performance across all three datasets.} This observation underscores the benefits of combining feature distribution with labels in enhancing model performance.

\begin{figure}[h!]
	\centering
	\includegraphics[width=1.0\linewidth]{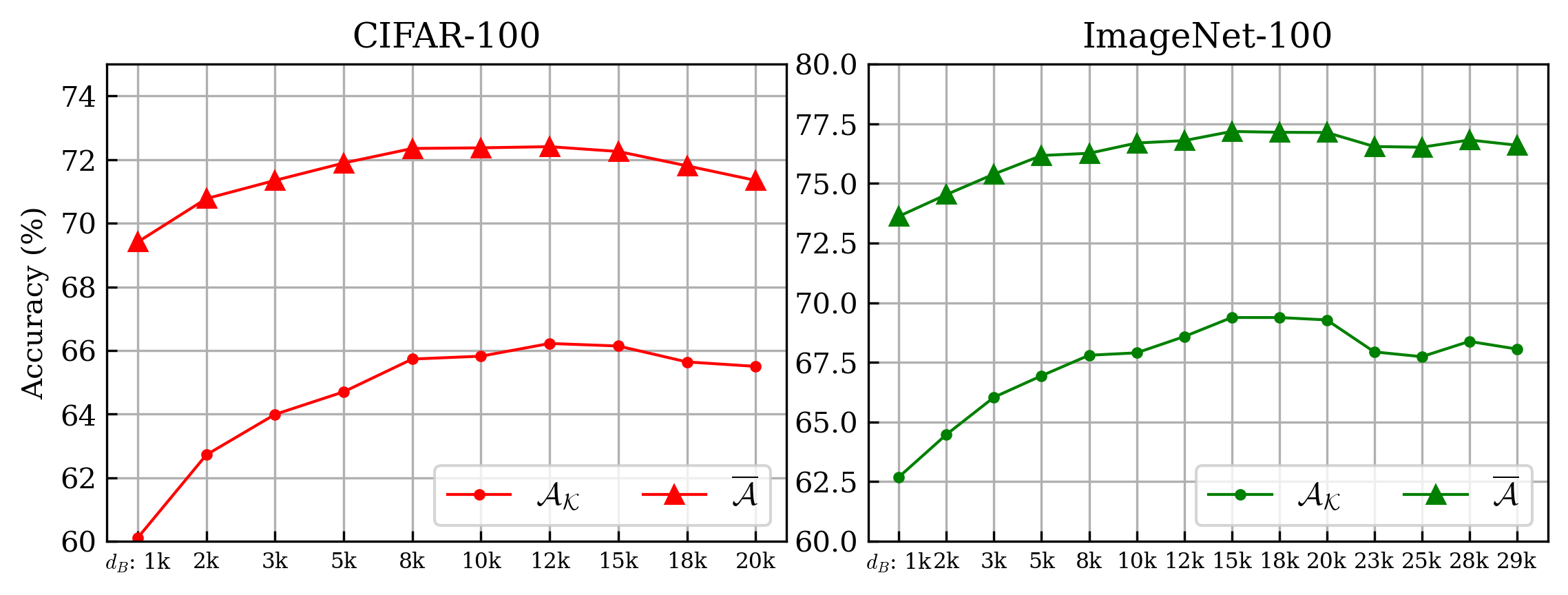}
	\caption{Accuracy on CIFAR-100 and ImageNet-100 with varying $d_{B}$.}
	\label{fig:acc_db}
\end{figure}

\textbf{Study on the Buffer Size of FFB.} To analyze the impact of buffer size $d_{B}$ on the FF module, we conduct the experiments with $d_{B} = \{1k, 2k,3k,5k,8k,10k,12k,15k,18k,20k\}$ on CIFAR-100 under the 5-phase setting. {We further extend the experiments of $d_{B} = \{23k, 25k, 28k, 29k\}$ on ImageNet-100 (a $d_{B}$ larger than $29k$ will {run out of memory})}. The results of average accuracy and last phase accuracy are depicted in Figure \ref{fig:acc_db}. On CIFAR-100, the performance of REAL first increases as $d_{B}$ scales up. The performance reaches the peak at $d_{B} = 12k$. {On ImageNet-100, a similar trend is observed as $d_{B}$ increases from $d_B = 1k$ to $d_B = 29k$ and the performance reaches the peak at $d_{B}=15k$.} This pattern shows that, the projection to relatively high dimensional space can boost the effect of {the FFB module and choices of $d_{B}=15k$ for both datasets yields good performance for REAL}.

\section{Conclusions}
\label{section:6}
\noindent  In this paper, we propose Representation Enhanced Analytic Learning (REAL) for EFCIL. To address the limitation of representation, REAL leverages a dual-stream base pretraining (DS-BPT) to initialize the network with base knowledge. {DS-BPT combines} a stream for general base knowledge acquisition via self-supervised contrastive learning (SSCL) and a stream for learning supervised feature distribution via supervised learning. The subsequent process of representation enhancing distillation (RED) in REAL enriches the SSCL pretrained backbones with additional knowledge obtained under supervision. {This process enhances the ability of the backbone to extract effective features at both base phase and unseen categories in future phases}, greatly facilitating subsequent stages of analytic class-incremental learning.{REAL addresses the underutilization of backbone knowledge by introducing the feature fusion buffer (FFB) to provide informative features.} FFB merges information from various backbone layers to create features with both transferable and specific information, thereby {enhancing} subsequent classifier training. Moreover, REAL can be seamlessly integrated into existing ACL techniques, including DS-AL. {Comprehensive experiments and analysis validate the effectiveness of REAL.} Empirical results showcase that REAL surpasses existing EFCIL methods and even outperforms most EBCIL methods.
\\

\noindent{\bf CRediT Authorship Contribution Statement}\\ 
	\indent Run He: Conceptualization, Investigation, Methodology, Implementation, Data Curation, Writing – original draft. Di Fang: Methodology, Implementation, Writing – review \& editing. Yizhu Chen: Methodology, Implementation, Writing – review \& editing. Kai Tong: Methodology, Writing – review \& editing. Cen Chen: Writing – review \& editing. Yi Wang: Writing – review \& editing. Lap-Pui Chau: Writing – review \& editing. Huiping Zhuang:  Conceptualization, Supervision, Writing – review \& editing.\\
	
	\noindent{\bf Declaration of Competing Interest}\\ 
	\indent The authors declare that they have no known competing financial interests or personal relationships that could have appeared to influence the work reported in this paper.\\
	
	\noindent{\bf Data Availability}\\ 
	\indent All datasets used in the paper are publicly available.\\
    
    \noindent{\bf Funding}\\ 
	\indent This work was supported in part by National Natural Science Foundation of China (Grant No. 62306117), in part by Guangzhou Basic and Applied Basic Research Foundation (Grant No. 2024A04J3681), and in part by GJYC Program of Guangzhou (Grant No. 2024D03J0005).
\bibliographystyle{elsarticle-num} 
\bibliography{references.bib}

@article{Online2023AAAI, 
	title   = {Online Hyperparameter Optimization for Class-Incremental Learning}, 
	author  = {Liu, Yaoyao and Li, Yingying and Schiele, Bernt and Sun, Qianru}, 
	journal = {Proceedings of the AAAI Conference on Artificial Intelligence}, 
	volume  = {37}, 
	number  = {7}, 
	year    = {2023}, 
	month   = {Jun.}, 
	pages   = {8906-8913},
	DOI     = {10.1609/aaai.v37i7.26070}, 
}

@InProceedings{FOSTER2022ECCV,
	author="Wang, Fu-Yun
	and Zhou, Da-Wei
	and Ye, Han-Jia
	and Zhan, De-Chuan",
	editor="Avidan, Shai
	and Brostow, Gabriel
	and Ciss{\'e}, Moustapha
	and Farinella, Giovanni Maria
	and Hassner, Tal",
	title="{FOSTER}: Feature Boosting and Compression for Class-Incremental Learning",
	booktitle="Computer Vision -- ECCV 2022",
	year="2022",
	publisher="Springer Nature Switzerland",
	address="Cham",
	pages="398--414",
	isbn="978-3-031-19806-9"
}

@InProceedings{SDC2020CVPR,
	author = {Yu, Lu and Twardowski, Bartlomiej and Liu, Xialei and Herranz, Luis and Wang, Kai and Cheng, Yongmei and Jui, Shangling and Weijer, Joost van de},
	title = {Semantic Drift Compensation for Class-Incremental Learning},
	booktitle = {Proceedings of the IEEE/CVF Conference on Computer Vision and Pattern Recognition (CVPR)},
	month = {June},
	year = {2020}
}

@InProceedings{resnet2016CVPR,
	author = {He, Kaiming and Zhang, Xiangyu and Ren, Shaoqing and Sun, Jian},
	title = {Deep Residual Learning for Image Recognition},
	booktitle = {Proceedings of the IEEE Conference on Computer Vision and Pattern Recognition (CVPR)},
	month = {June},
	year = {2016}
}

@article{ImageNet2015IJCV,
	title={{ImageNet} large scale visual recognition challenge},
	author={Russakovsky, Olga and Deng, Jia and Su, Hao and Krause, Jonathan and Satheesh, Sanjeev and Ma, Sean and Huang, Zhiheng and Karpathy, Andrej and Khosla, Aditya and Bernstein, Michael and others},
	journal={International Journal of Computer Vision},
	volume={115},
	number={3},
	pages={211--252},
	year={2015},
	publisher={Springer}
}

@article{CIFAR2009,
  title={Learning multiple layers of features from tiny images},
  author={Krizhevsky, Alex and Hinton, Geoffrey and others},
  year={2009},
  publisher={Toronto, ON, Canada}
}

@article{cil_review2021NNs,
	author = {Eden Belouadah and Adrian Popescu and Ioannis Kanellos},
	title = {A comprehensive study of class incremental learning algorithms for visual tasks},
	journal = {Neural Networks},
	volume = {135},
	pages = {38-54},
	year = {2021}
}

@article{pil2001,
	title={Pseudoinverse learning algorithm for feedforward neural networks},
	author={Guo, Ping and Lyu, Michael R and Mastorakis, NE},
	journal={Advances in Neural Networks and Applications},
	pages={321-326},
	year={2001},
	publisher={Puerto De La Cruz, Tenerife, Canary Islands, Spain}
}

@article{cpnet2021,
	title={Correlation Projection for Analytic Learning of a Classification Network},
	author={Zhuang, Huiping and Lin, Zhiping and Toh, Kar-Ann},
	journal={Neural Processing Letters},
	pages={1--22},
	year={2021},
	publisher={Springer}
}

@article{brmp2021,
	author={Zhuang, Huiping and Lin, Zhiping and Toh, Kar-Ann},
	journal={IEEE Transactions on Systems, Man, and Cybernetics: Systems}, 
	title={Blockwise Recursive {Moore-Penrose} Inverse for Network Learning}, 
	year={2021},
	volume={},
	number={},
	pages={1-14}
}

@InProceedings{LUCIR2019_CVPR,
	author = {Hou, Saihui and Pan, Xinyu and Loy, Chen Change and Wang, Zilei and Lin, Dahua},
	title = {Learning a Unified Classifier Incrementally via Rebalancing},
	booktitle = {Proceedings of the IEEE/CVF Conference on Computer Vision and Pattern Recognition (CVPR)},
	month = {June},
	year = {2019}
}

@article{EWC2017nas,
	title={Overcoming catastrophic forgetting in neural networks},
	author={Kirkpatrick, James and Pascanu, Razvan and Rabinowitz, Neil and Veness, Joel and Desjardins, Guillaume and Rusu, Andrei A and Milan, Kieran and Quan, John and Ramalho, Tiago and Grabska-Barwinska, Agnieszka and others},
	journal={Proceedings of the national academy of sciences},
	volume={114},
	number={13},
	pages={3521--3526},
	year={2017},
	publisher={National Acad Sciences}
}

@article{LwF2018TPAMI,
	
	author={Li, Zhizhong and Hoiem, Derek},
	
	journal={IEEE Transactions on Pattern Analysis and Machine Intelligence}, 
	
	title={Learning without Forgetting}, 
	
	year={2018},
	
	volume={40},
	
	number={12},
	
	pages={2935-2947}
}

@InProceedings{iCaRL2017_CVPR,
	author = {Rebuffi, Sylvestre-Alvise and Kolesnikov, Alexander and Sperl, Georg and Lampert, Christoph H.},
	title = {{iCaRL}: Incremental Classifier and Representation Learning},
	booktitle = {Proceedings of the IEEE Conference on Computer Vision and Pattern Recognition (CVPR)},
	month = {July},
	year = {2017}
}

@article{RMM2021NeuriPS,
	title={{RMM}: Reinforced Memory Management for Class-Incremental Learning},
	author={Liu, Yaoyao and Schiele, Bernt and Sun, Qianru},
	journal={Advances in Neural Information Processing Systems},
	volume={34},
	year={2021}
}

@InProceedings{AANet_2021_CVPR,
	author    = {Liu, Yaoyao and Schiele, Bernt and Sun, Qianru},
	title     = {Adaptive Aggregation Networks for Class-Incremental Learning},
	booktitle = {Proceedings of the IEEE/CVF Conference on Computer Vision and Pattern Recognition (CVPR)},
	month     = {June},
	year      = {2021},
	pages     = {2544-2553}
}

@InProceedings{Mnemonics_2020_CVPR,
	author = {Liu, Yaoyao and Su, Yuting and Liu, An-An and Schiele, Bernt and Sun, Qianru},
	title = {Mnemonics Training: Multi-Class Incremental Learning Without Forgetting},
	booktitle = {Proceedings of the IEEE/CVF Conference on Computer Vision and Pattern Recognition (CVPR)},
	month = {June},
	year = {2020}
}

@inproceedings{podnet2020ECCV,
	title={{PodNet}: Pooled outputs distillation for small-tasks incremental learning},
	author={Douillard, Arthur and Cord, Matthieu and Ollion, Charles and Robert, Thomas and Valle, Eduardo},
	booktitle={Computer Vision--ECCV 2020: 16th European Conference, Glasgow, UK, August 23--28, 2020, Proceedings, Part XX 16},
	pages={86--102},
	year={2020},
	organization={Springer}
}

@inproceedings{ACIL2022NeurIPS,
	author = {Zhuang, Huiping and Weng, Zhenyu and Wei, Hongxin and XIE, RENCHUNZI and Toh, Kar-Ann and Lin, Zhiping},
	booktitle = {Advances in Neural Information Processing Systems},
	editor = {S. Koyejo and S. Mohamed and A. Agarwal and D. Belgrave and K. Cho and A. Oh},
	pages = {11602--11614},
	publisher = {Curran Associates, Inc.},
	title = {{ACIL}: Analytic Class-Incremental Learning with Absolute Memorization and Privacy Protection},
	volume = {35},
	year = {2022}
}

@InProceedings{FeTrIL2023WACV,
	author    = {Petit, Gr\'egoire and Popescu, Adrian and Schindler, Hugo and Picard, David and Delezoide, Bertrand},
	title     = {{FeTrIL}: Feature Translation for Exemplar-Free Class-Incremental Learning},
	booktitle = {Proceedings of the IEEE/CVF Winter Conference on Applications of Computer Vision (WACV)},
	month     = {January},
	year      = {2023},
	pages     = {3911-3920}
}

@InProceedings{SSRE2022CVPR,
	author    = {Zhu, Kai and Zhai, Wei and Cao, Yang and Luo, Jiebo and Zha, Zheng-Jun},
	title     = {Self-Sustaining Representation Expansion for Non-Exemplar Class-Incremental Learning},
	booktitle = {Proceedings of the IEEE/CVF Conference on Computer Vision and Pattern Recognition (CVPR)},
	month     = {June},
	year      = {2022},
	pages     = {9296-9305}
}

@InProceedings{PASS2021CVPR,
	author    = {Zhu, Fei and Zhang, Xu-Yao and Wang, Chuang and Yin, Fei and Liu, Cheng-Lin},
	title     = {Prototype Augmentation and Self-Supervision for Incremental Learning},
	booktitle = {Proceedings of the IEEE/CVF Conference on Computer Vision and Pattern Recognition (CVPR)},
	month     = {June},
	year      = {2021},
	pages     = {5871-5880}
}

@InProceedings{SimSiam2021CVPR,
	author    = {Chen, Xinlei and He, Kaiming},
	title     = {Exploring Simple Siamese Representation Learning},
	booktitle = {Proceedings of the IEEE/CVF Conference on Computer Vision and Pattern Recognition (CVPR)},
	month     = {June},
	year      = {2021},
	pages     = {15750-15758}
}

@InProceedings{GKEAL2023CVPR,
	author    = {Zhuang, Huiping and Weng, Zhenyu and He, Run and Lin, Zhiping and Zeng, Ziqian},
	title     = {{GKEAL}: Gaussian Kernel Embedded Analytic Learning for Few-Shot Class Incremental Task},
	booktitle = {Proceedings of the IEEE/CVF Conference on Computer Vision and Pattern Recognition (CVPR)},
	month     = {June},
	year      = {2023},
	pages     = {7746-7755}
}

@ARTICLE{SSL2023TKDE,
  author={Liu, Xiao and Zhang, Fanjin and Hou, Zhenyu and Mian, Li and Wang, Zhaoyu and Zhang, Jing and Tang, Jie},
  journal={IEEE Transactions on Knowledge and Data Engineering}, 
  title={Self-Supervised Learning: Generative or Contrastive}, 
  year={2023},
  volume={35},
  number={1},
  pages={857-876},
  doi={10.1109/TKDE.2021.3090866}}

@InProceedings{SimCLR2020ICML,
  title = 	 {A Simple Framework for Contrastive Learning of Visual Representations},
  author =       {Chen, Ting and Kornblith, Simon and Norouzi, Mohammad and Hinton, Geoffrey},
  booktitle = 	 {Proceedings of the 37th International Conference on Machine Learning},
  pages = 	 {1597--1607},
  year = 	 {2020},
  editor = 	 {III, Hal Daumé and Singh, Aarti},
  volume = 	 {119},
  series = 	 {Proceedings of Machine Learning Research},
  month = 	 {13--18 Jul},
  publisher =    {PMLR}}

@InProceedings{MOCO2020CVPR,
author = {He, Kaiming and Fan, Haoqi and Wu, Yuxin and Xie, Saining and Girshick, Ross},
title = {Momentum Contrast for Unsupervised Visual Representation Learning},
booktitle = {Proceedings of the IEEE/CVF Conference on Computer Vision and Pattern Recognition (CVPR)},
month = {June},
year = {2020}
}

@inproceedings{BYOL2020NeurIPS,
 author = {Grill, Jean-Bastien and Strub, Florian and Altch\'{e}, Florent and Tallec, Corentin and Richemond, Pierre and Buchatskaya, Elena and Doersch, Carl and Avila Pires, Bernardo and Guo, Zhaohan and Gheshlaghi Azar, Mohammad and Piot, Bilal and kavukcuoglu, koray and Munos, Remi and Valko, Michal},
 booktitle = {Advances in Neural Information Processing Systems},
 editor = {H. Larochelle and M. Ranzato and R. Hadsell and M.F. Balcan and H. Lin},
 pages = {21271--21284},
 publisher = {Curran Associates, Inc.},
 title = {Bootstrap Your Own Latent - A New Approach to Self-Supervised Learning},
 volume = {33},
 year = {2020}
}

@ARTICLE{SSL2020TPAMI,
  author={Jing, Longlong and Tian, Yingli},
  journal={IEEE Transactions on Pattern Analysis and Machine Intelligence}, 
  title={Self-Supervised Visual Feature Learning With Deep Neural Networks: A Survey}, 
  year={2021},
  volume={43},
  number={11},
  pages={4037-4058},
  doi={10.1109/TPAMI.2020.2992393}}

@InProceedings{PRAKA2023ICCV,
    author    = {Shi, Wuxuan and Ye, Mang},
    title     = {Prototype Reminiscence and Augmented Asymmetric Knowledge Aggregation for Non-Exemplar Class-Incremental Learning},
    booktitle = {Proceedings of the IEEE/CVF International Conference on Computer Vision (ICCV)},
    month     = {October},
    year      = {2023},
    pages     = {1772-1781}
}

@article{AAAI2024DSAL, 
title={{DS-AL}: A Dual-Stream Analytic Learning for Exemplar-Free Class-Incremental Learning}, 
volume={38}, 
DOI={10.1609/aaai.v38i15.29670}, 
number={15}, 
journal={Proceedings of the AAAI Conference on Artificial Intelligence}, 
author={Zhuang, Huiping and He, Run and Tong, Kai and Zeng, Ziqian and Chen, Cen and Lin, Zhiping}, 
year={2024}, 
month={Mar.},
pages={17237-17244} }

@InProceedings{NAPA-VQ2023ICCV,
    author    = {Malepathirana, Tamasha and Senanayake, Damith and Halgamuge, Saman},
    title     = {{NAPA-VQ}: Neighborhood-Aware Prototype Augmentation with Vector Quantization for Continual Learning},
    booktitle = {Proceedings of the IEEE/CVF International Conference on Computer Vision (ICCV)},
    month     = {October},
    year      = {2023},
    pages     = {11674-11684}
}

@InProceedings{TPCIL2020ECCV,
author="Tao, Xiaoyu
and Chang, Xinyuan
and Hong, Xiaopeng
and Wei, Xing
and Gong, Yihong",
editor="Vedaldi, Andrea
and Bischof, Horst
and Brox, Thomas
and Frahm, Jan-Michael",
title="Topology-Preserving Class-Incremental Learning",
booktitle="Computer Vision -- ECCV 2020",
year="2020",
publisher="Springer International Publishing",
address="Cham",
pages="254--270",
isbn="978-3-030-58529-7"
}

@InProceedings{GeoDL2021CVPR,
    author    = {Simon, Christian and Koniusz, Piotr and Harandi, Mehrtash},
    title     = {On Learning the Geodesic Path for Incremental Learning},
    booktitle = {Proceedings of the IEEE/CVF Conference on Computer Vision and Pattern Recognition (CVPR)},
    month     = {June},
    year      = {2021},
    pages     = {1591-1600}
}

@InProceedings{SI2017ICML,
  title = 	 {Continual Learning Through Synaptic Intelligence},
  author =       {Friedemann Zenke and Ben Poole and Surya Ganguli},
  booktitle = 	 {Proceedings of the 34th International Conference on Machine Learning},
  pages = 	 {3987--3995},
  year = 	 {2017},
  editor = 	 {Precup, Doina and Teh, Yee Whye},
  volume = 	 {70},
  series = 	 {Proceedings of Machine Learning Research},
  month = 	 {06--11 Aug},
  publisher =    {PMLR}
}

@InProceedings{MAS2018ECCV,
author="Aljundi, Rahaf
and Babiloni, Francesca
and Elhoseiny, Mohamed
and Rohrbach, Marcus
and Tuytelaars, Tinne",
editor="Ferrari, Vittorio
and Hebert, Martial
and Sminchisescu, Cristian
and Weiss, Yair",
title="Memory Aware Synapses: Learning What (not) to Forget",
booktitle="Computer Vision -- ECCV 2018",
year="2018",
publisher="Springer International Publishing",
address="Cham",
pages="144--161",
isbn="978-3-030-01219-9"
}

@InProceedings{PC2018ICML,
  title = 	 {{{Progress and Compress}}: A scalable framework for continual learning},
  author =       {Schwarz, Jonathan and Czarnecki, Wojciech and Luketina, Jelena and Grabska-Barwinska, Agnieszka and Teh, Yee Whye and Pascanu, Razvan and Hadsell, Raia},
  booktitle = 	 {Proceedings of the 35th International Conference on Machine Learning},
  pages = 	 {4528--4537},
  year = 	 {2018},
  editor = 	 {Dy, Jennifer and Krause, Andreas},
  volume = 	 {80},
  series = 	 {Proceedings of Machine Learning Research},
  month = 	 {10--15 Jul},
  publisher =    {PMLR}
}

@ARTICLE{CRNet2023TPAMI,
  author={Li, Depeng and Zeng, Zhigang},
  journal={IEEE Transactions on Pattern Analysis and Machine Intelligence}, 
  title={{CRNet}: A Fast Continual Learning Framework With Random Theory}, 
  year={2023},
  volume={45},
  number={9},
  pages={10731-10744},
  doi={10.1109/TPAMI.2023.3262853}}

@InProceedings{RWalk2018ECCV,
author="Chaudhry, Arslan
and Dokania, Puneet K.
and Ajanthan, Thalaiyasingam
and Torr, Philip H. S.",
editor="Ferrari, Vittorio
and Hebert, Martial
and Sminchisescu, Cristian
and Weiss, Yair",
title="Riemannian Walk for Incremental Learning: Understanding Forgetting and Intransigence",
booktitle="Computer Vision -- ECCV 2018",
year="2018",
publisher="Springer International Publishing",
address="Cham",
pages="556--572",
isbn="978-3-030-01252-6"
}

@InProceedings{LWM2019CVPR,
author = {Dhar, Prithviraj and Singh, Rajat Vikram and Peng, Kuan-Chuan and Wu, Ziyan and Chellappa, Rama},
title = {Learning Without Memorizing},
booktitle = {Proceedings of the IEEE/CVF Conference on Computer Vision and Pattern Recognition (CVPR)},
month = {June},
year = {2019}
}

@InProceedings{Fusion2022CVPR,
    author    = {Toldo, Marco and Ozay, Mete},
    title     = {Bring Evanescent Representations to Life in Lifelong Class Incremental Learning},
    booktitle = {Proceedings of the IEEE/CVF Conference on Computer Vision and Pattern Recognition (CVPR)},
    month     = {June},
    year      = {2022},
    pages     = {16732-16741}
}

@inproceedings{POLO2023MM,
author = {Wang, Shaokun and Shi, Weiwei and He, Yuhang and Yu, Yifan and Gong, Yihong},
title = {Non-Exemplar Class-Incremental Learning via Adaptive Old Class Reconstruction},
year = {2023},
isbn = {9798400701085},
publisher = {Association for Computing Machinery},
address = {New York, NY, USA},
doi = {10.1145/3581783.3611926},
booktitle = {Proceedings of the 31st ACM International Conference on Multimedia},
pages = {4524–4534},
numpages = {11},
location = {, Ottawa ON, Canada, },
series = {MM '23}
}

@article{tSNE2008JMLR,
  title={Visualizing data using {t-SNE}},
  author={Van der Maaten, Laurens and Hinton, Geoffrey},
  journal={Journal of machine learning research},
  volume={9},
  number={11},
  year={2008}
}

@InProceedings{GDumb2020ECCV,
author="Prabhu, Ameya
and Torr, Philip H. S.
and Dokania, Puneet K.",
editor="Vedaldi, Andrea
and Bischof, Horst
and Brox, Thomas
and Frahm, Jan-Michael",
title="{GDumb}: A Simple Approach that Questions Our Progress in Continual Learning",
booktitle="Computer Vision -- ECCV 2020",
year="2020",
publisher="Springer International Publishing",
address="Cham",
pages="524--540",
isbn="978-3-030-58536-5"
}

@inproceedings{MIR2019NeurIPS,
 author = {Aljundi, Rahaf and Belilovsky, Eugene and Tuytelaars, Tinne and Charlin, Laurent and Caccia, Massimo and Lin, Min and Page-Caccia, Lucas},
 booktitle = {Advances in Neural Information Processing Systems},
 editor = {H. Wallach and H. Larochelle and A. Beygelzimer and F. d\textquotesingle Alch\'{e}-Buc and E. Fox and R. Garnett},
 pages = {},
 publisher = {Curran Associates, Inc.},
 title = {Online Continual Learning with Maximal Interfered Retrieval},
 volume = {32},
 year = {2019}
}

@ARTICLE{Cover1965TEC,
  author={Cover, Thomas M.},
  journal={IEEE Transactions on Electronic Computers}, 
  title={Geometrical and Statistical Properties of Systems of Linear Inequalities with Applications in Pattern Recognition}, 
  year={1965},
  volume={EC-14},
  number={3},
  pages={326-334},
  doi={10.1109/PGEC.1965.264137}}

@inproceedings{RAIL2024NeurIPS,
  title     = {Advancing Cross-domain Discriminability in Continual Learning of Vision-Language Models},
  author    = {Yicheng Xu and Yuxin Chen and Jiahao Nie and Yusong Wang and Huiping Zhuang and Manabu Okumura},
  booktitle = {The Thirty-eighth Annual Conference on Neural Information Processing Systems},
  year      = {2024},
  no_url    = {https://openreview.net/forum?id=boGxvYWZEq}
}

@inproceedings{MMAL2024MM,
  author    = {Yue, Xianghu and Zhang, Xueyi and Chen, Yiming and Zhang, Chengwei and Lao, Mingrui and Zhuang, Huiping and Qian, Xinyuan and Li, Haizhou},
  title     = {{MMAL}: Multi-Modal Analytic Learning for Exemplar-Free Audio-Visual Class Incremental Tasks},
  year      = {2024},
  isbn      = {9798400706868},
  publisher = {Association for Computing Machinery},
  address   = {New York, NY, USA},
  no_url    = {https://doi.org/10.1145/3664647.3681607},
  doi       = {10.1145/3664647.3681607},
  booktitle = {Proceedings of the 32nd ACM International Conference on Multimedia},
  pages     = {2428--2437},
  numpages  = {10},
  location  = {Melbourne VIC, Australia},
  series    = {MM '24}
}

@inproceedings{GACL2024NeurIPS,
 author = {Zhuang, Huiping and Chen, Yizhu and Fang, Di and He, Run and Tong, Kai and Wei, Hongxin and Zeng, Ziqian and Chen, Cen},
 booktitle = {Advances in Neural Information Processing Systems},
 editor = {A. Globerson and L. Mackey and D. Belgrave and A. Fan and U. Paquet and J. Tomczak and C. Zhang},
 pages = {83024--83047},
 publisher = {Curran Associates, Inc.},
 title = {{GACL}: Exemplar-Free Generalized Analytic Continual Learning},
 volume = {37},
 year = {2024}
}

@misc{AIR2024,
    title={{AIR}: Analytic Imbalance Rectifier for Continual Learning}, 
    author={Di Fang and Yinan Zhu and Zhiping Lin and Cen Chen and Ziqian Zeng and Huiping Zhuang},
    year={2024},
    eprint={2408.10349},
    archivePrefix={arXiv},
    primaryClass={cs.LG},
    no_url={https://arxiv.org/abs/2408.10349}, 
}

@article{pycil,
    author = {Da-Wei Zhou and Fu-Yun Wang and Han-Jia Ye and De-Chuan Zhan},
    title = {{PyCIL}: a Python toolbox for class-incremental learning},
    journal = {Science China Information Sciences},
    year = {2023},
    volume = {66},
    number = {9},
    pages = {197101},
    doi = {10.1007/s11432-022-3600-y}
}

@inproceedings{FeCAM2023NeurIPS,
    author = {Goswami, Dipam and Liu, Yuyang and Twardowski, Bart\l omiej and van de Weijer, Joost},
    booktitle = {Advances in Neural Information Processing Systems},
    no_editor = {A. Oh and T. Naumann and A. Globerson and K. Saenko and M. Hardt and S. Levine},
    pages = {6582--6595},
    publisher = {Curran Associates, Inc.},
    title = {{FeCAM}: Exploiting the Heterogeneity of Class Distributions in Exemplar-Free Continual Learning},
    no_url = {https://proceedings.neurips.cc/paper_files/paper/2023/file/15294ba2dcfb4521274f7aa1c26f4dd4-Paper-Conference.pdf},
    volume = {36},
    year = {2023}
}

@InProceedings{CaSSLe2022CVPR,
    author    = {Fini, Enrico and da Costa, Victor G. Turrisi and Alameda-Pineda, Xavier and Ricci, Elisa and Alahari, Karteek and Mairal, Julien},
    title     = {Self-Supervised Models Are Continual Learners},
    booktitle = {Proceedings of the IEEE/CVF Conference on Computer Vision and Pattern Recognition (CVPR)},
    month     = {June},
    year      = {2022},
    pages     = {9621-9630}
}

@InProceedings{Co2L2021ICCV,
    author    = {Cha, Hyuntak and Lee, Jaeho and Shin, Jinwoo},
    title     = {{Co2L}: Contrastive Continual Learning},
    booktitle = {Proceedings of the IEEE/CVF International Conference on Computer Vision (ICCV)},
    month     = {October},
    year      = {2021},
    pages     = {9516-9525}
}

@InProceedings{SSLEFCIL2024WACV,
    author    = {Petit, Gr\'egoire and Soumm, Micha\"el and Feillet, Eva and Popescu, Adrian and Delezoide, Bertrand and Picard, David and Hudelot, C\'eline},
    title     = {An Analysis of Initial Training Strategies for Exemplar-Free Class-Incremental Learning},
    booktitle = {Proceedings of the IEEE/CVF Winter Conference on Applications of Computer Vision (WACV)},
    month     = {January},
    year      = {2024},
    pages     = {1837-1847}
}

@inproceedings{FOAL2024NeurIPS,
 author = {Zhuang, Huiping and Liu, Yuchen and He, Run and Tong, Kai and Zeng, Ziqian and Chen, Cen and Wang, Yi and Chau, Lap-Pui},
 booktitle = {Advances in Neural Information Processing Systems},
 editor = {A. Globerson and L. Mackey and D. Belgrave and A. Fan and U. Paquet and J. Tomczak and C. Zhang},
 pages = {41517--41538},
 publisher = {Curran Associates, Inc.},
 title = {{F-OAL}: Forward-only Online Analytic Learning with Fast Training and Low Memory Footprint in Class Incremental Learning},
 volume = {37},
 year = {2024}
}

@inproceedings{NIPS2014_shallow_deep_feature,
 author = {Yosinski, Jason and Clune, Jeff and Bengio, Yoshua and Lipson, Hod},
 booktitle = {Advances in Neural Information Processing Systems},
 editor = {Z. Ghahramani and M. Welling and C. Cortes and N. Lawrence and K.Q. Weinberger},
 pages = {},
 publisher = {Curran Associates, Inc.},
 title = {How transferable are features in deep neural networks?},
 volume = {27},
 year = {2014}
}

@article{SASS2023KBS,
title = {Semantic alignment with self-supervision for class incremental learning},
journal = {Knowledge-Based Systems},
volume = {282},
pages = {111114},
year = {2023},
issn = {0950-7051},
doi = {10.1016/j.knosys.2023.111114},
author = {Zhiling Fu and Zhe Wang and Xinlei Xu and Mengping Yang and Ziqiu Chi and Weichao Ding}
}

@article{SAGG2024KBS,
title = {Sharpness-aware gradient guidance for few-shot class-incremental learning},
journal = {Knowledge-Based Systems},
volume = {299},
pages = {112030},
year = {2024},
issn = {0950-7051},
doi = {10.1016/j.knosys.2024.112030},
author = {Runhang Chen and Xiao-Yuan Jing and Fei Wu and Haowen Chen}
}

@article{KGP2025KBS,
title = {Knowledge-guided prompt-based continual learning: Aligning task-prompts through contrastive hard negatives},
journal = {Knowledge-Based Systems},
volume = {310},
pages = {113009},
year = {2025},
issn = {0950-7051},
doi = {10.1016/j.knosys.2025.113009},
author = {Heng-yang Lu and Long-kang Lin and Chenyou Fan and Chongjun Wang and Wei Fang and Xiao-jun Wu}
}

@article{AEF-OCLTVT2025,
  author    = {Zhuang, Huiping and Fang, Di and Tong, Kai and Liu, Yuchen and Zeng, Ziqian and Zhou, Xu and Chen, Cen},
  journal   = {IEEE Transactions on Vehicular Technology}, 
  title     = {Online Analytic Exemplar-Free Continual Learning With Large Models for Imbalanced Autonomous Driving Task}, 
  year      = {2025},
  volume    = {74},
  number    = {2},
  pages     = {1949--1958},
  doi       = {10.1109/TVT.2024.3483557},
  ISSN      = {1939-9359},
  month     = feb,
}

@inproceedings{SupCon2020NeurIPS,
 author = {Khosla, Prannay and Teterwak, Piotr and Wang, Chen and Sarna, Aaron and Tian, Yonglong and Isola, Phillip and Maschinot, Aaron and Liu, Ce and Krishnan, Dilip},
 booktitle = {Advances in Neural Information Processing Systems},
 editor = {H. Larochelle and M. Ranzato and R. Hadsell and M.F. Balcan and H. Lin},
 pages = {18661--18673},
 publisher = {Curran Associates, Inc.},
 title = {Supervised Contrastive Learning},
 url = {https://proceedings.neurips.cc/paper_files/paper/2020/file/d89a66c7c80a29b1bdbab0f2a1a94af8-Paper.pdf},
 volume = {33},
 year = {2020}
}

@InProceedings{Comp2021ICML,
  title = 	 {Compositional Few-Shot Class-Incremental Learning},
  author =       {Zou, Yixiong and Zhang, Shanghang and Zhou, Haichen and Li, Yuhua and Li, Ruixuan},
  booktitle = 	 {Proceedings of the 41st International Conference on Machine Learning},
  pages = 	 {62964--62977},
  year = 	 {2024},
  editor = 	 {Salakhutdinov, Ruslan and Kolter, Zico and Heller, Katherine and Weller, Adrian and Oliver, Nuria and Scarlett, Jonathan and Berkenkamp, Felix},
  volume = 	 {235},
  series = 	 {Proceedings of Machine Learning Research},
  month = 	 {21--27 Jul},
  publisher =    {PMLR},
  url = 	 {https://proceedings.mlr.press/v235/zou24c.html}
}

@inproceedings{
ViT,
title={An Image is Worth 16x16 Words: Transformers for Image Recognition at Scale},
author={Alexey Dosovitskiy and Lucas Beyer and Alexander Kolesnikov and Dirk Weissenborn and Xiaohua Zhai and Thomas Unterthiner and Mostafa Dehghani and Matthias Minderer and Georg Heigold and Sylvain Gelly and Jakob Uszkoreit and Neil Houlsby},
booktitle={International Conference on Learning Representations},
year={2021},
url={https://openreview.net/forum?id=YicbFdNTTy}
}






\end{document}